\documentclass{tlp}

\usepackage{chemarrow}
\usepackage{url}

  \title[CIFF Proof Procedure for ALP with constraints]{The CIFF Proof Procedure for Abductive Logic Programming with Constraints: Theory, Implementation and
  Experiments}

  \author[P. Mancarella et al.]
         {Paolo Mancarella and Giacomo Terreni\\
           Dipartimento di Informatica, Universit\`a di Pisa \\
           \email{paolo.mancarella@unipi.it}\\
           \email{terreni@di.unipi.it}
          \and Fariba Sadri and Francesca Toni\\
           Department of Computing, Imperial College London \\
           \email{fs@doc.ic.ac.uk}\\
           \email{ft@doc.ic.ac.uk}
          \and Ulle Endriss\\
         Institute for Logic, Language \& Computation (ILLC), University of Amsterdam\\
         \email{ulle.endriss@uva.nl}
         }

\newtheorem{definition}{Definition}[section]
\newtheorem{proposition}{Proposition}[section]
\newtheorem{theorem}{Theorem}[section]
\newtheorem{example}{Example}[section]
\newtheorem{corollary}{Corollary}[section]
\newtheorem{lemma}{Lemma}[section]

\newcommand{\sequenza}[2]{\ensuremath{#1_1, \ldots, #1_{#2}}}

\newcommand{\NAF}{\ensuremath{\neg}}
\newcommand{\AnswerC}{$\langle\Delta, \sigma, \Gamma\rangle$}
\newcommand{\answerC}{\ensuremath{\langle \Gamma, E, DE \rangle}}

\newcommand{\IC}{\ensuremath{IC}}
\newcommand{\Abd}{\ensuremath{A}}
\newcommand{\Prog}{\ensuremath{P}}

\newcommand{\Asys}{\ensuremath{{\cal A}}-System}
\newcommand{\apici}[1]{``#1''}
\newcommand{\CIFF}{CIFF System 4.0}

\newcommand{\abdciffframe}{\ensuremath{\langle Th, A, IC \rangle_{\Re}}}

\newcommand{\answerciff}{\ensuremath{\langle \Delta, C \rangle}}
\newcommand{\answerciffC}{\ensuremath{\langle \Gamma, E, DE \rangle}}
\newcommand{\answerciffCcup}{\ensuremath{\Gamma \cup E \cup DE}}

\newcommand{\Der}{\ensuremath{{\cal D}}}
\newcommand{\Branch}{\ensuremath{{\cal B}}}

\newcommand{\imp}{\rightarrow}

\newcommand{\ruledeff}[4]{%
\begin{quote}\noindent\begin{tabular}{ll}
\multicolumn{2}{l}{\textbf{#1}} \\
\hline
\textbf{Given:}  &  #2 \\
\textbf{Conditions:}  &  #3 \\
\textbf{Action:} &  #4 \\ \hline
\end{tabular}
\end{quote}}

\newcommand{\ruledef}[5]{%
\begin{quote}\noindent\begin{tabular}{ll}
\multicolumn{2}{l}{\textbf{R#1 - #2}} \\ \hline
\textbf{Given:}  & \{ #3 \}\\
\textbf{Conditions:}  & \{ #4 \}\\
\textbf{Action:} &  #5 \\ \hline
\end{tabular}
\end{quote}}

\newcommand{\ruledeflong}[6]{%
\begin{quote}\noindent\begin{tabular}{ll}
\multicolumn{2}{l}{\textbf{R#1 - #2}} \\ \hline
\textbf{Given:}  & \{ #3 \} \\
\textbf{Conditions:}  & \{ #4  \\
  & \ #5 \} \\
\textbf{Action:} &  #6 \\ \hline
\end{tabular}\end{quote}
}

\newcommand{\ruleapp}{\ensuremath{F \: \autorightarrow{$N$,$\Cs$}{$\phi$} \: F'}}
\newcommand{\ruleappargs}[4]{\ensuremath{#1 \: \autorightarrow{#2}{#3} \: #4}}
\newcommand{\ruleappder}{\ensuremath{F_i \: \autorightarrow{$N_i,\Cs_i$}{$\phi_i$} \: F_{i+1}}}

\newcommand{\equivalent}{\leftrightarrow}

\newcommand{\univquant}{universally quantified}
\newcommand{\existquant}{existentially quantified}
\newcommand{\Th}{\mbox{\it Th}}
\newcommand{\CIFFNaf}{{CIFF\ensuremath{^{\neg}}}}

\newcommand{\CIFFSys}{CIFF System}

\newcommand{\CIFFQ}{CIFF 4.0}

\newcommand{\Abdprog}{\ensuremath{\langle P, \: A, \: IC\rangle}}
\newcommand{\AbdCprog}{\ensuremath{\langle P, \: A, \: IC\rangle_{\Re}}}
\newcommand{\CIFFprog}{$\langle\Th, \: \Abd, \: \IC\rangle_{\Re}$}

\newcommand{\Answer}{$\langle\Delta,\sigma\rangle$}

\newcommand{\Nodes}{\ensuremath{\cal{N}}}
\newcommand{\Cs}{\ensuremath{\chi}}
\newcommand{\Csp}{\ensuremath{\Psi}}
\newcommand{\Sel}{\ensuremath{{\cal S}}}
\newcommand{\SelA}[1]{$\cal{S}$$(#1)$}

\newcommand{\Undef}{\mbox{\it undefined}}
\newcommand{\RewEqs}[1]{$\cal{E}$\ensuremath{(#1)}}

\newcommand{\replace}{\textbf{replace}}
\newcommand{\replaceall}{\textbf{replace\_all}}
\newcommand{\add}{\textbf{add}}
\newcommand{\delete}{\textbf{delete}}
\newcommand{\marc}{\textbf{mark}}

\renewcommand{\bot}{false}
\renewcommand{\top}{true}

\def\eqq{\doteq}

\submitted{23 January 2008}
\revised{23 December 2008}
\accepted{22 April 2009}

\begin{document}
\maketitle


\begin{abstract}
\noindent We present the CIFF proof procedure for abductive logic
programming with constraints, and we prove its correctness. CIFF is
an extension of the IFF proof procedure for abductive logic
programming, relaxing the original restrictions over variable
quantification (\emph{allowedness conditions}) and incorporating a
constraint solver to deal with numerical constraints as in
constraint logic programming. Finally, we describe the \CIFFSys,
comparing it with state of the art abductive systems and answer set
solvers and showing how to use it to program some applications.\\
(To appear in Theory and Practice of Logic Programming - TPLP).

\end{abstract}


\section{Introduction}
\label{intro} Abduction has found broad application as a powerful
tool for hypothetical reasoning with incomplete knowledge. This form
of reasoning is handled by labeling some pieces of information as
abducibles, i.e. as possible hypotheses, that can be assumed to
hold, provided that they are consistent with the rest of the given
information in the knowledge base.

\noindent Attempts to make abductive reasoning an effective
computational tool have given rise to \emph{Abductive Logic
Programming} (ALP) which combines abduction with standard logic
programming. A number of \emph{abductive proof procedures} have been
proposed in the literature, e.g.
\cite{km90,km91,console91,SLDNFA97,iff97}. These differ in that they
rely upon different semantics, the most common being the
(generalized) stable models semantics \cite{km90} and the
(three-valued) completion semantics \cite{3comp}. Many of these
proof procedures enrich the expressive power of the abductive
framework by allowing the inclusion of \emph{integrity constraints}
(ICs) to further restrict the range of possible hypotheses.

\noindent ALP has also been integrated with \emph{Constraint Logic
Programming} (CLP) \cite{CLP,jaffar98}, in order to combine
abductive reasoning with an arithmetic tool for {\em constraint
solving} \cite{Asystem1,ACLP-JLP00,ALPCfirst2,abdchr2} (in the sense
of CLP, not to be confused with integrity constraints). In recent
years, several proof procedures for ALP with constraints (ALPC) have
been proposed, including ACLP~\cite{ACLP-JLP00} and the \Asys{}
\cite{Asystem1}.

\noindent Important applications of ALP and ALPC include agent
programming
 ~\cite{KGP,kgp04,JELIA02-francesca}, (semantic) web
management applications~\cite{toniinfo}, planning and combinatorial
problems \cite{ALPCfirst1,ALPCfirst2}.

\medskip
\noindent Here we propose CIFF, another proof procedure for ALPC
which extends the IFF procedure \cite{iff97} in two ways, namely (1)
by integrating abductive reasoning with constraint solving, and (2)
by relaxing the allowedness conditions on suitable inputs given in
\cite{iff97}, in order to be able to handle a wider class of
problems. The CIFF proof procedure has been implemented in Prolog in
the CIFF System \cite{CIFFManual}.

\noindent CIFF features have been exploited in various application
domains. In \cite{KGP,kgp04} CIFF has been used as the computational
core for modelling an agent's planning, reactivity and temporal
reasoning capabilities based on a variant of the abductive event
calculus \cite{EC,Shanahan89}. Also, a (slightly modified) prototype
version of CIFF  for checking and repairing XML web sites is
currently under development \cite{WWWCIF,WWV08CIF,CIFFThesis}.

\medskip
\noindent We have compared empirically the \CIFFSys{} to other
related systems, namely the \Asys{} \cite{Asystem1,Asystemthesis},
which is the closest system from both a theoretical and an
implementative viewpoint, and two state-of-the-art answer set
solvers: SMODELS \cite{smodels1,smodels2} and DLV \cite{dlv1,dlv2}.
These solvers implement a different (answer set) semantics
\cite{AnswerSets}, but share with our approach the objective of
modeling dynamic and non-monotonic settings in a declarative (and
thus human-oriented) way. The results of our tests show that (1) the
\CIFFSys{} and the other systems have comparable performances and
(2) the \CIFFSys{} is able to handle 
variables taking values in unbound domains.\medskip

\noindent The paper is organised as follows. In the next section we
give background notions about ALPC. Section \ref{sec:ciffmain}
specifies the CIFF proof procedure, while formal results are shown
in Section \ref{sec:soundness}. In Section \ref{system} we briefly
describe the CIFF System and in Section \ref{cap4:secComparison} we
discuss some related work together with some experimental results.
Finally, Section \ref{cap4:secConclusions} concludes the paper and
proposes some future work.

\medskip
\noindent This paper combines and extends a number of earlier
papers: \cite{ciffJELIA}, defining an earlier version of the CIFF
proof procedure, \cite{CIFFSystem}, \cite{CIFFARW} and
\cite{LPNMRCIF} all defining earlier versions of the \CIFFSys{}.

\section{Abductive Logic Programming with Constraints}
\label{back}

\noindent We present here some background 
on ALPC. We will assume familiarity with basic concepts  of Logic
Programming (atom, term etc.) as found e.g. in \cite{LP}. We will
frequently write $\vec{t}$ for a vector of terms such as
$t_1,\ldots,t_k$. For instance, we are going to write $p(\vec{t})$
rather than $p(t_1,\ldots,t_k)$. Throughout the paper, to simplify
the presentation, we will assume that predicates cannot have the
same name but different arities. Moreover, with an abuse of
notation, we will often use disjunctions and conjunctions as if they
were sets, and similarly for substitutions. In particular, we will
abstract away from the position of a conjunct (respectively
disjunct) in a conjunction (respectively disjunction) and we will
apply to disjunctions and conjunctions set-theoretic operations such
as union, inclusion, difference and so on.

\medskip
\noindent An \emph{abductive logic program} is a tuple \Abdprog{}
where:

\begin{itemize}
    \item \Prog{} is a \emph{normal logic program}, namely a set of \emph{clauses} of the
    form:
\[p(\vec{s}) \leftarrow l_1(\vec{t}_1) \wedge \ldots \wedge l_n(\vec{t}_n) \qquad n \geq 0\]
where $p(\vec{s})$ is an atom and each $l_i(\vec{t}_i)$ is a
literal, i.e. an atom $a(\vec{t})$ or the negation of an atom
$a(\vec{t})$, represented as $\neg a(\vec{t})$. We refer to
$p(\vec{s})$ as the \emph{head} of the clause and to $l_1(\vec{t}_1)
\wedge \ldots \wedge l_n(\vec{t}_n)$ as the \emph{body} of the
clause. A predicate $p$ occurring in the head of at least one clause
in $P$ is called a \emph{defined predicate} and the set of clauses
in $P$ such that $p$ occurs in their heads is called the
\emph{definition set} of $p$.

Any variable in a clause is implicitly universally quantified with
scope the entire clause.

\item \Abd{} is a set of predicates, referred to as \emph{abducible predicates}. Atoms whose
predicate is an abducible predicate are referred to as
\emph{abducible atoms} or simply as \emph{abducibles}. Abducible
atoms must not occur in the head of any clause of \Prog{} (without
loss of generality, see \cite{abdsurvey98}).

\item \IC{} is a set of \emph{integrity constraints} which are \emph{implications} of the form:

\[l_1(\vec{t}_1) \wedge \ldots \wedge l_n(\vec{t}_n) \;\imp\; a_1(\vec{s}_1) \vee \ldots \vee a_m(\vec{s}_m) \qquad n,m \geq 0 \quad n + m \geq 1\]

Each of the $l_i(\vec{t}_i)$ is a literal (as defined above) while
each of the $a_i(\vec{s}_i)$ is an atom.
We refer to $l_1(\vec{t}_1) \wedge \ldots \wedge l_n(\vec{t}_n)$ as the \emph{body} 
and to $a_1(\vec{s}_1) \vee \ldots \vee a_m(\vec{s}_m)$ as the
\emph{head} of the integrity constraint.

Any variable in an integrity constraint is implicitly universally
quantified with scope the entire implication.
\end{itemize}

\noindent Given an abductive logic program \Abdprog, we will refer
to the set of all (defined and abducible) predicates occurring in
\Abdprog{} as its {\em Herbrand signature}. Moreover, as is the
convention in LP, we will assume as given a {\em Herbrand universe},
namely a set of ground terms.
Further, we will refer to
all ground atoms whose predicate belongs to the {\em Herbrand
signature} of \Abdprog{} and that can be built using terms
in the Herbrand universe as the {\em Herbrand base}
of \Abdprog. Finally, we will refer to {\em Herbrand terms} as
(ground and non ground) terms whose instances belong to the Herbrand universe.
Then, a \emph{query} $Q$ to an abductive logic program \Abdprog{} is
a conjunction of literals whose predicate belongs to the Herbrand
signature of \Abdprog{} and whose arguments are Herbrand terms.
Any variable occurring in
$Q$ is implicitly existentially quantified with scope $Q$. 

\medskip
\noindent A normal logic program $\Prog$ provides definitions for
certain predicates,
while abducibles can be 
used to extend these definitions to form possible
\emph{explanations} for queries, which can be regarded as
\emph{observations} against the background of the world knowledge
encoded in the given abductive logic program. Integrity constraints,
on the other hand, restrict the range of possible explanations. Note
that, in general, the set of abducible predicates may not coincide
with the set of all predicates without definitions in \Prog{} (i.e.
the set of \emph{open} predicates).

\noindent Informally, given an abductive logic program \Abdprog{}
and a query $Q$, an explanation for a query $Q$ is a set of (ground)
abducible atoms $\Delta$ that, together with \Prog, both ``entails''
(an appropriate ground instantiation of) $Q$, with respect to some
notion of ``entailment'', and ``satisfies'' the set of integrity
constraints \IC{} (see \cite{abdsurvey98} for possible notions of
integrity constraint ``satisfaction'').
%
The notion of ``entailment'' depends on the semantics associated
with the logic program \Prog{} (there are many different possible
choices for such semantics \cite{abdsurvey98}).

\noindent The following definition of abductive answer formalizes
this informal notion of explanation.

\begin{definition}[Abductive answer]\label{def:abdanswer} An \emph{abductive
answer} to a query $Q$ with respect to an abductive logic program
\Abdprog{} is a pair \Answer, where $\Delta$ is a finite set of
ground abducible atoms and $\sigma$ is a ground substitution for the
(existentially quantified) variables occurring in $Q$, such that:

\begin{itemize}
    \item $\Prog\cup \Delta \models_{LP} Q\sigma$ and
    \item $\Prog\cup \Delta \models_{LP} IC$
\end{itemize}

\noindent where $\models_{LP}$ stands for the chosen semantics for
logic programming. 
\end{definition}

\noindent Given an abductive logic program \Abdprog, an {abductive
answer} to a query $Q$ provides an {explanation} for $Q$, understood
as an observation: the answer specifies which instances of the
abducible predicates have to be assumed to hold for the
(corresponding instances of the) observation $Q$ to hold as well,
and, in addition, it forces such an explanation to validate the
integrity constraints.

\medskip
\noindent The framework of abductive logic programming can be
usefully extended to handle constraint predicates in the same way
Constraint Logic Programming (CLP) \cite{CLP} extends logic
programming. The CLP framework is defined over a particular
structure $\Re$ consisting of a domain $D(\Re)$, and a set of
constraint predicates which includes equality ($\eqq$) and disequality
($\neq$), together with an assignment of relations on $D(\Re)$ for
each constraint predicate. We will refer to the set of constraint
predicates in $\Re$ as the {\em constraint signature} (of $\Re$),
and to atoms of the constraint predicates as {\em constraint atoms}
(over $\Re$).

\noindent The structure $\Re$ is equipped with a notion of
$\Re$-satisfiability. Given a set of (possibly non-ground)
constraint atoms $C$, the fact that $C$ is $\Re$-satisfiable will be
denoted as $\models_{\Re} C$. Moreover we denote as $\sigma
\models_{\Re} C$ the fact that the grounding $\sigma$ of the
variables of $C$ over $D(\Re)$ satisfies $C$, i.e. $C$ is
$\Re$-satisfied.

\medskip
\noindent An \emph{abductive logic program with constraints} is a
tuple \AbdCprog{} with all components defined as above but where
constraint atoms for $\Re$ might occur in the body of clauses of
\Prog{} and of integrity constraints of \IC. Also, queries for
abductive logic programs with constraints might include constraint
atoms (over $\Re$). We keep the notion of Herbrand signature and
Herbrand base as before.

\medskip
\noindent The semantics of CLP is obtained by combining the logic
programming semantics $\models_{LP}$ and the notion of
$\Re$-satisfiability \cite{CLP}. We denote this semantic notion as
$\models_{LP(\Re)}$ and we use it in the notion of abductive answer
with respect to an abductive logic program with constraints.

\begin{definition}[Abductive answer with constraints]\label{def:abdcanswer}
An \emph{abductive answer with constraints} to a query $Q$ with
respect to an abductive logic program with constraints \AbdCprog{}
is a tuple \AnswerC, where $\Delta$ is a finite set of abducible
atoms, $\sigma$ is a ground substitution for the (existentially
quantified) variables occurring
in $Q$ and $\Gamma$ is a set of constraint atoms 
such that

\begin{enumerate}
  \item there exists a ground substitution $\sigma'$ for the variables occurring in $\Gamma\sigma$ such that $\sigma' \models_{\Re} \Gamma\sigma$
and
  \item for each ground substitution $\sigma'$ for the variables occurring in $\Gamma\sigma$ such that $\sigma' \models_{\Re} \Gamma\sigma$,
   there exists a ground substitution $\sigma''$ for the variables
occurring in $Q \cup \Delta \cup \Gamma$, with $\sigma\sigma'
\subseteq \sigma''$, such that:

\begin{itemize}

    \item $\Prog \cup \Delta\sigma'' \models_{LP(\Re)} Q\sigma''$ and
    \item $\Prog \cup \Delta\sigma'' \models_{LP(\Re)} IC.$
\end{itemize}
\end{enumerate}

\end{definition}

\medskip

\begin{example}\label{ex:example1} Consider the following abductive logic
program with constraints (here we assume that $<$ is a constraint
predicate of $\Re$ with the expected semantics):

\[\begin{array}{l@{\quad}l}
 \Prog: & p(X) \leftarrow q(T_1,T_2) \wedge T_1\!\!<\!X \wedge X\!\!<\!8 \\
        & q(X_1,X_2) \leftarrow s(X_1,a) \\[1pt]
 \Abd: & \{ r, s\}\\[1pt]
 \IC: & r(Z) \;\imp\; p(Z)\\[1pt]
\end{array}\]

\noindent An {abductive answer with constraints} for the query $Q =
r(6)$ is

\[ \langle \{ r(6), s(T_1,a) \}, \oslash, \{ T_1 < 6\} \rangle \]

\noindent
where $\oslash$ is the empty set.

\noindent Intuitively, given the query $r(6)$, the integrity
constraint in $\IC$ would fire and force the atom $p(6)$ to hold,
which in turn requires $s(T_1,a)$ for some $T_1<6$ to be true.

\medskip
\noindent Considering a non-ground version of the query, for example
$Q = r(Y)$, the following is an abductive answer with constraints:

\[ \langle \{ r(Y), s(T_1,a) \}, \{ Y/5 \}, \{ T_1 < Y, Y < 8\} \rangle. \]

\end{example}

\section{The CIFF Proof Procedure}\label{sec:ciffmain}
 \noindent The language of CIFF is the same of an
abductive logic program with constraints, but we assume to have the
special symbols $\bot$ and $\top$.
%
These will be used, in particular, to represent the empty
body ($\top$) and the empty head ($\bot$) of an integrity
constraint.

\noindent The CIFF framework relies upon the availability of a
concrete CLP structure $\Re$ over arithmetical domains equipped at
least with the set $\{ <, \leq,
>, \geq, \eqq, \neq\}$ of constraint predicates whose intended
semantics is the expected one\footnote{Here $\eqq$ is used for equality instead of $=$,
the latter being used to stand for Clark's equality as shown later.}.
The set of constraint predicates is
assumed to be closed under complement\footnote{Clearly, $\neq$ is the complement of $\eqq$ and viceversa.}.
When needed, we will denote
by $\overline{Con}$ the complement of the constraint atom $Con$
(e.g. $\overline{X < 3}$ is $X \geq 3$). We also assume that the
constraint domain offers a set of functions like $+, -, * \ldots$
whose semantics is again the expected one.

\noindent The structure $\Re$ is a \emph{black box} component in the
definition of the CIFF proof procedure: for handling constraint
atoms and evaluating constraint functions, we rely upon an
underlying \emph{constraint solver} over $\Re$ which is assumed to
be both
 sound and complete with respect to $\models_{\Re}$. In particular we will assume
 that, given a constraint atom $Con$ and its complement
 $\overline{Con}$, the formulae $Con \vee \overline{Con}$ and  $Con \rightarrow
 Con$ are tautologies with respect to the constraint solver
 semantics. We do not commit
 to any concrete implementation of a constraint solver, hence the range of the admissible
 arguments to constraint predicates ($D(\Re)$) depends on
the specifics of the chosen constraint solver. 

\medskip
\noindent The semantics of the CIFF proof procedure is defined in
terms of Definition \ref{def:abdcanswer} where (1) the constraint
structure $\Re$ is defined as above, and (2) the semantics of logic
programming is the three-valued completion semantics \cite{3comp}
(we denote as $\models_{3(\Re)}$ the notion of $\models_{LP(\Re)}$
with respect to that semantics). We refer to an abductive answer
with constraints as a \emph{CIFF abductive answer}. Recall that the
three-valued completion semantics embeds the Clark Equality Theory
\cite{Clark1978}, denoted by \emph{CET}, which handles equalities
over Herbrand terms.

\medskip
\noindent The CIFF proof procedure operates on a set of
\emph{iff-definitions} obtained from the \emph{completion}
\cite{Clark1978} of the defined predicates $\sequenza{p}{n}$ in the
Herbrand signature of \AbdCprog{}.

\noindent The completion of a predicate $p$ with respect to
\AbdCprog{} is defined as follows.
  Assume that the following set of clauses is the definition set of
  $p$ in \AbdCprog:

\begin{center}
\begin{tabular}{llr}
$p(\vec{t}_1)$ & $\leftarrow$ & $D_1$ \\
 & $\vdots$  & \\
$p(\vec{t}_k)$ &  $\leftarrow$ & $D_k$ \\
\end{tabular}
\end{center}

\noindent where each $D_i$ is a conjunction of literals and
constraint atoms. The \emph{iff-definition} of $p$ is of the form:

\[p(\vec{X}) \;\equivalent\; [\vec{X} = \vec{t}_1 \wedge D_1]\vee\cdots\vee [\vec{X} = \vec{t}_k \wedge D_k]\]

\noindent where $\vec{X}$ is a vector of \emph{fresh} variables (not
occurring in any $D_i$ or $t_i$) implicitly \emph{universally}
quantified with scope the entire iff-definition, and all other
variables are implicitly \emph{existentially} quantified with scope
 the right-hand side disjunct in which it occurs.

\noindent
Note that the equality symbol $=$ is used to represent Clark's equality in
\emph{iff-definitions}. In the sequel, we will refer to $=$ as the \emph{equality predicate}
and to atoms containing it as \emph{equality atoms}\footnote{In particular, constraints of the form
$A \eqq B$ are \emph{not} equality atoms but they are (equality) constraint atoms.}. Note also that
input programs can not include $=$ explicitly, $=$ being reserved for Clark's equality in
\emph{iff-definitions}.


\noindent If $p$ is a non-abducible, non-constraint, non-equality
atom and it does not occur in the head of any clause of $P$ its
iff-definition is of the form:

\[p(\vec{X}) \;\equivalent\; \bot.\]

\begin{definition}[CIFF Theory and CIFF Framework]
Let \AbdCprog{} be an abductive logic program with constraints. The
\emph{CIFF theory} $Th$ relative to \AbdCprog{} is the set of all
the iff-definitions of each non-abducible, non-constraint predicate
 in the language of \AbdCprog{}. Moreover we say that a \emph{CIFF
framework} is the tuple \CIFFprog.

\end{definition}

\begin{example}\label{iffdef}
Let us consider the following abductive logic program with constraints \AbdCprog:

\[\begin{array}{l@{\quad}l}
\Prog: & p(T) \leftarrow s(T) \\
 & p(W) \leftarrow W\!\!<\!8 \\[1pt]
\Abd: & \{ s\}\\[1pt]
\IC: & r(T) \wedge s(T)  \;\imp\; p(T)\\[1pt]
\end{array}\]

\noindent The resulting CIFF theory $Th$ is:\medskip

\[\begin{array}{ll}
p(X) & \leftrightarrow [X = T \wedge s(T)] \vee [X = W \wedge W\!\!<\!8] \\[1pt]
r(Y) & \leftrightarrow \bot. \\[1pt]
\end{array}\]

\noindent With explicit quantification, the theory $Th$ would be:

\[\begin{array}{lll}
\forall X & (p(X) & \leftrightarrow [\exists T (X = T \wedge s(T)] \vee [\exists W (X = W \wedge W\!\!<\!8)]) \\[1pt]
\forall Y & (r(Y) & \leftrightarrow \bot). \\[1pt]
\end{array}\]

\noindent Note that $Th$ includes an iff-definition for $r$ even
though $r$ occurs only in the integrity constraints \IC. Moreover,
there is no iff-definition for the abducible predicate $s$. To
improve readability and unless otherwise stated, in the remainder we
will write CIFF theories with implicit variable quantification.

\end{example}

\begin{definition}[CIFF query]
A \emph{CIFF query} $Q$ is a conjunction of literals, possibly
including constraint literals. All the variables in a CIFF query $Q$
are implicitly existentially quantified with scope $Q$.
\end{definition}

\paragraph{Allowedness.}
\citeN{iff97} require frameworks for their IFF proof procedure to
meet a number of so-called \emph{allowedness conditions} to be able
to guarantee the correct operation of their proof procedure. These
conditions are designed to avoid problematic patterns of
quantification which can lead to problems analogous to \emph{floundering} in
LP with negation \cite{LP}. These allowedness conditions are primarily
needed to avoid dealing with atomic conjuncts which may
contain {\em universally} quantified variables, and also to avoid keeping
explicit quantifiers for the variables which are introduced during an IFF
computation.

\noindent
Informally, the problem arises when a universally
quantified variable occurring in a clause occurs nowhere else in the body
except, possibly, in a negative literal or in an abducible atom.

\noindent The IFF proof procedure for abductive logic programming
(without constraints) has the following allowedness conditions:

\begin{itemize}
\item an integrity constraint $A\imp B$ is allowed iff every variable in it
also occurs in an atomic conjunct within its body $A$;
\item an iff-definition $p(\vec{X})\equivalent D_1\vee\cdots\vee D_n$ is
allowed iff every variable, other than those in $\vec{X}$, occurring
 in a disjunct $D_i$, also occurs
inside a non-equality atomic conjunct within the same $D_i$;
\item a query is allowed iff every variable in it also occurs in an atomic conjunct within the query
itself.
\end{itemize}

\noindent As stated in \cite{iff97}, the above allowedness conditions
ensure statically that floundering is avoided. We will refer to a CIFF framework
 arising from an abductive logic program without constraints and to a query
 $Q$ such that they
 are allowed as above as \emph{IFF allowed}.

\noindent Also our CIFF frameworks \CIFFprog{} must be
\emph{allowed} in order to guarantee the correct operation of CIFF.
Unfortunately, it is difficult to formulate appropriate allowedness
conditions that guarantee correct execution of the proof procedure
without imposing too many unnecessary restrictions. This is a
well-known problem, which is further aggravated for languages that
include constraint predicates. In particular, adapting the IFF
approach, the allowedness condition for an iff-definition would be
defined as follows:

\begin{definition}[CIFF Static Allowedness]\label{def:StaticAllowedness}
\noindent A CIFF framework \abdciffframe{} is \emph{CIFF-statically
allowed} iff it satisfies the following conditions:

\begin{itemize}
\item each integrity constraint $A\imp B \in IC$ is such that every variable in it
also occurs in a non-constraint atomic conjunct within
its body $A$;
\item each iff-definition $p(\vec{X})\equivalent D_1\vee\cdots\vee D_n \in Th$ is
such that every variable, other than those in $\vec{X}$, occurring
 in a disjunct $D_i$, also occurs
in a non-equality, non-constraint atomic conjunct within the same
$D_i$.
\end{itemize}
\noindent A CIFF query $Q$ is \emph{CIFF-statically allowed} iff every
variable in $Q$ also occurs in a non-constraint atomic conjunct within the query
itself.
\end{definition}

\noindent Our proposal is to relax the above allowedness
conditions, and to check \emph{dynamically}, i.e.\ at
runtime, the risk of floundering. Some restrictions are still needed in order
to ensure that the quantification of variables during a CIFF computation
can be kept implicit, both for simplicity and for keeping the IFF style of
behaviour.

\noindent
The new allowedness conditions for CIFF are defined as follows.

\begin{definition}[CIFF Allowedness]\label{def:CIFFallowed}
\noindent A CIFF framework \CIFFprog{} is \emph{CIFF-allowed} iff every
iff-definition in \Th{} is allowed. An iff-definition
$p(\vec{X})\equivalent D_1\vee\cdots\vee D_n$ is allowed iff every
variable, other than those in $\vec{X}$, occurring
 in a disjunct $D_i$, also occurs
inside an atomic conjunct within the same $D_i$.

\noindent A CIFF query $Q$ is \emph{CIFF-allowed} iff every variable in
it also occurs in an atomic conjunct within the query itself.
\end{definition}

\noindent Note that in this definition there are no restrictions
concerning the integrity constraints. Moreover, it is worth noting
that for a query $Q$, the notions of IFF allowedness, CIFF static
allowedness and CIFF allowedness for $Q$ are identical.

\begin{example}
The following CIFF framework is CIFF allowed ($\Prog_1$ is the
original normal logic program):

\[\begin{array}{l@{\quad}l}
\Prog_1: & p(Z)\\[1pt]
     & p(Y) \leftarrow \NAF q(Y)\\[1pt]
\Th_1: & p(X) \leftrightarrow [X=Z] \vee [X=Y \wedge \NAF q(Y)]\\[1pt]
& q(X) \leftrightarrow \bot\\[1pt]
& s(X) \leftrightarrow \bot\\[1pt]
\Abd_1: & \oslash\\[1pt]
\IC_1: & Z=W   \;\imp\; s(Z,W)\\
\end{array}\]

\noindent It is worth noting that the above CIFF framework is neither
CIFF statically allowed nor IFF allowed (note that there are no
constraints in it). Indeed, in $\Th_1$, the variable $Z$ occurs only
in an \emph{equality atomic} conjunct and the variable $Y$ occurs
only in an \emph{equality atomic} conjunct and in a negative
literal. The following CIFF framework, instead, is not CIFF allowed
 (\Prog$_2$ is the original normal logic program):

\[\begin{array}{l@{\quad}l}
\Prog_2: & p(Z) \leftarrow \NAF q(Z,Y)\\[1pt]
\Th_2: & p(X) \leftrightarrow [X=Z \wedge \NAF q(Z,Y)]\\[1pt]
& q(X,Y) \leftrightarrow \bot\\[1pt]
& s(X,Y) \leftrightarrow \bot\\[1pt]
\Abd_2: & \oslash\\[1pt]
\IC_2: & q(Z, W) \;\imp\; s(Z,W)\\[1pt]
\end{array}\]

\noindent The non-allowedness is due to the variable $Y$ in $Th_2$
 which occurs only in a negative literal.

\noindent The query $Q = \NAF q(V, a)$ is not CIFF allowed (and it is
neither CIFF statically allowed nor IFF allowed) due to the variable
$V$ which occurs only in a negative literal.

Note that in some cases a CIFF framework which is not CIFF allowed can be turned
into a CIFF allowed framework by adding explicit, though useless since trivially satisfied,
constraints over the {\em critical} variables (e.g. $Y$ in $\Th_2$ above). For instance, the above non CIFF-allowed framework
can be modified by changing the first clause as follows:

\[\begin{array}{l@{\quad}l}
\Prog_2: & p(Z) \leftarrow \NAF q(Z,Y) \wedge Y \eqq Y\\[1pt]
\end{array}\]
\end{example}

\noindent
Note however that this can be done only if
the critical variables such as $Y$ above are
meant to be variables ranging over the domain $D(\Re)$, i.e. they are constraint variables.

\medskip
\noindent The following example shows how the IFF allowedness
requirement forbids the use of the IFF proof procedure even for
simple abductive frameworks where IFF could compute correct
abductive answers.

\begin{example}
Consider the following CIFF framework:

\[\begin{array}{l@{\quad}l}
\Prog_3: & p(Y).\\
&   q(Z) \leftrightarrow r(Z) \wedge p(a)\\[2pt]
\Th_3: & p(X) \leftrightarrow [X=Y]\\
&   q(X) \leftrightarrow [X = Z \wedge r(Z) \wedge p(a)]\\[2pt]
\Abd_3: & \{ r \}\\[1pt]
\IC_3: & \oslash
\end{array}\]

\noindent The above framework is not IFF allowed due to the
variable $Y$. Consider the query $q(b)$. Intuitively there is a
simple and sound abductive answer for $q(b)$, i.e. $r(b)$ and this
could be computed by IFF, were it not for the allowedness
restrictions it imposes on its inputs. Instead, the above framework
is CIFF allowed and, as will become clear, the CIFF proof procedure returns exactly the
correct answer.
\end{example}

\noindent Until now we have shown only \apici{artificial} examples,
but the IFF allowedness restrictions limit the
application of the IFF proof procedure in many realistic settings. 

\begin{example}
Abduction is a very interesting solution for modeling agent systems
and agent capabilities. In particular the Abductive Event Calculus
(AEC) language \cite{AEC1,AEC2} is a popular framework for modeling
(among others) planning capabilities of an agent through abductive
reasoning. The following CIFF framework models a fragment of the
AEC (definitions for $init$ and $term$ are omitted for simplicity).

\[\begin{array}{ll@{\quad}l}
AEC: & holds(G,T) \leftarrow & happens(A,T_1) \wedge init(A,G) \wedge \\
& & \NAF clip(T_1,G,T) \wedge T_1 < T\\
& clip(T_1,G,T_2) \leftarrow & happens(A,T) \wedge term(A,G) \wedge T_1 \leq T \wedge T < T_2\\[3pt]
\Th_{AEC}: & holds(X_1,X_2) \leftrightarrow & [X_1 = G \wedge X_2 = T
\wedge happens(A,T_1) \wedge \\ & & init(A,G) \wedge \NAF clip(T_1,G,T) \wedge T_1 < T]\\[1pt]
 &   clip(X_1,X_2,X_3) \leftrightarrow & [X_1 = T_1 \wedge X_2 = G
\wedge X_3 = T_2\  \wedge \\ & & happens(A,T) \wedge
        term(A,G) \wedge T_1 \leq T \wedge T < T_2]\\[3pt]
\Abd_{AEC}: & \{ happens \}&\\[1pt]
\IC_{AEC}: & \oslash &
\end{array}\]

\noindent The above framework is neither an IFF framework due to the
presence of constraint atoms, nor CIFF statically allowed due to the
variable $T$ in the first iff-definition and the variables $T_1$ and
$T_2$ in the second iff-definition, violating the allowedness
restrictions stated in Definition \ref{def:StaticAllowedness}. This
is because these variables occur only in equality and/or constraint
atomic conjuncts in the respective disjuncts. However the framework
is CIFF allowed and CIFF can be used for reasoning with it, as done,
e.g., in the KGP model \cite{KGP}.
\end{example}

\noindent In the remainder of the paper, we will always assume that
CIFF frameworks and CIFF queries are CIFF allowed. For simplicity,
from here onwards, with the word \emph{allowed} we mean \emph{CIFF
allowed}, unless otherwise explicitly stated.
\subsection{CIFF Proof Rules} \label{sec:ciffrules} \noindent The
CIFF proof procedure is a rewriting procedure, consisting of a
number of \emph{CIFF proof rules}, each of which replaces a
\emph{CIFF formula} by another one.

\noindent In the remainder, a negative literal $L = \NAF A$,
everywhere in a CIFF framework, in a CIFF query, or in a CIFF
formula, will be written in implicative form, i.e. $\NAF A$  is
written as  $A \;\imp\; \bot$.

\noindent Hence, in this context a literal is either an atom $A$ or
an implication $A \rightarrow \bot$.

\noindent A special case of such implication is given by the next
definition.

\begin{definition}[CIFF Disequality]\label{def:ciffdiseq} A \emph{CIFF disequality} is an implication of the form

\begin{center}$X = t \rightarrow \bot$\end{center}

\noindent where $X$ is an \emph{existentially quantified variable}
and $t$ is a term \emph{not in the form of a universally quantified
variable} and such that $X$ does not occur in $t$.
\end{definition}


\begin{definition}[CIFF formula, CIFF node and CIFF conjunct]\label{def:node} A \emph{CIFF formula} $F$ is a disjunction

\[ N_1 \vee \ldots \vee N_n \qquad n \geq 0. \]

\noindent If $n=0$, the disjunction is equivalent to $\bot$.

\noindent Each disjunct $N_i$ is a \emph{CIFF node} which is of the
form:

\[ C_1 \wedge \ldots \wedge C_m \qquad m \geq 0. \]

\noindent If $m=0$, the conjunction is equivalent to $\top$. Each
conjunct $C_i$ is a \emph{CIFF conjunct} and it can be of the form
of:
\begin{itemize}
    \item an atom (\emph{atomic CIFF conjunct}),
    \item an implication (\emph{implicative CIFF conjunct}, including negative literals) or
    \item a disjunction of conjunctions of literals (\emph{disjunctive
    CIFF conjunct})
\end{itemize}

\noindent where implications are of the form:

\[ L_1 \wedge \ldots \wedge L_t \rightarrow A_1 \vee \ldots \vee A_s \qquad s,t \geq 1, \]

\noindent where each $L_i$ is a literal (possibly $\bot$ or $\top$)
and each $A_i$ is an atom (possibly $\bot$ or $\top$).

\noindent In the sequel we will refer to $L_1 \wedge \ldots \wedge L_t$ as the \emph{body}
of the implication and to $A_1 \vee \ldots \vee A_s$ as the \emph{head} of the implication.

\noindent In a \emph{CIFF node} $N$, variables which appear either
in an atomic CIFF conjunct or in a disjunctive CIFF conjunct are
implicitly existentially quantified with scope $N$. All the
remaining variables, i.e. variables occurring only in implicative
CIFF conjuncts, are implicitly universally quantified with the scope
being the implication in which they appear.

\medskip
\noindent Finally a CIFF node $N$ can have an associated
\emph{label} $\lambda$. We will denote a node $N$ labeled by
$\lambda$ as $\lambda : N$. 
\end{definition}

\noindent We are now going to present the \emph{CIFF proof rules}.
In doing that, we treat a CIFF node as a \emph{(multi)set} of CIFF
conjuncts and a CIFF formula as a \emph{(multi)set} of CIFF nodes.
I.e. we represent a CIFF formula $F = N_1 \vee \ldots \vee N_n$ as

\[ \{ N_1, \ldots, N_n\} \]

\noindent where each $N_i$ is a CIFF node, of the form $C_1 \wedge
\ldots \wedge C_m$ represented by

\[ \{ C_1, \ldots, C_m\} \]

\noindent where each $C_j$ is a CIFF conjunct.

\begin{example}
\noindent Let us consider the following abductive logic program with
constraints \AbdCprog:

\[\begin{array}{l@{\quad}l}
\Prog: & p \leftarrow a \\[1pt]
 & p \leftarrow b \\[1pt]
\Abd: & \{ a,b,c\}\\[1pt]
\IC: & a \rightarrow c\\[1pt]
\end{array}\]

\noindent The CIFF formula $p \wedge (a \rightarrow c)$ (composed of
a single node) is represented by:

\[ \{ \{ p, (a \rightarrow c) \} \} \]


\noindent The CIFF formula $[a \wedge (a \rightarrow c)] \vee [b
\wedge (a \rightarrow c)]$, composed of two CIFF nodes (obtained in CIFF
from the earlier nodes as will be seen later) $N_1 = a
\wedge (a \rightarrow c)$ and $N_2 = b \wedge (a \rightarrow c)$ is
represented by:

\[ \{ \{ a, (a \rightarrow c) \}, \{ b, (a \rightarrow c) \}  \}. \]
\end{example}

\noindent Each \emph{CIFF proof rule}\footnote{In the remainder,
when we want to refer to a CIFF framework, a CIFF node, a CIFF
formula and so on, we drop the prefix \apici{CIFF} if it is clear
from the context.} operates over a node $N$ within a formula $F$ and
it will result in a new formula $F'$. A rule is presented in the
following form:

\ruledeff{Rule name $\phi$ \qquad Input: $F,N$ \qquad Output $F'$}{a
set of CIFF conjuncts \Cs{} in $N$}{a set of conditions over \Cs{}
and $N$}
    {\{\replace, \replaceall, \add, \delete\} \Csp ; \marc $\: \lambda$ }


\noindent The \textbf{Given} part identifies a (possibly empty) set
of conjuncts \Cs{} in $N$ within $F$. A rule $\phi$ can be applied
on a set \Cs{} of conjuncts of $N$ satisfying the stated
\textbf{Conditions}. We say $\phi$ is \emph{applicable} to $F$ and
we call the set \Cs{} a \emph{rule input} for $\phi$. Finally, the
\textbf{Action} part defines both a new set of conjuncts \Csp{} and
an action (\replace, \replaceall, \add, \delete{} or \marc) which
states, as described below, how $F'$ is obtained from $F$ through
\Csp{}. In the remainder we will omit to specify the \textbf{Input}
part and the \textbf{Output} part.

\noindent Given a rule $\phi$ as above, we denote by

\begin{center}
 \ruleapp
\end{center}

\noindent the \emph{application} of rule $\phi$ with \textbf{Input}
$F,N$, \textbf{Given} $\Cs$, and \textbf{Output} $F'$.

 \noindent Abstracting from the particular
action, $F'$ is always derived from $F$ replacing the node $N$ by a
set of nodes \Nodes, i.e.:

    \[F' = F -  \{N\}  \cup \Nodes \]

\noindent We refer to \Nodes{} as the \emph{CIFF successor nodes} of
$N$ and we refer to each node $N' \in \Nodes$ as a \emph{CIFF
successor node} of $N$. Each type of action defines \Nodes{} as
follows:

\medskip
\noindent\begin{tabular}{ll}
    \replace: &  \Nodes $=  \{ (N - \Cs)  \cup  \Csp \}$\\
    \replaceall: & \Nodes $= \{[(N - \Cs) \cup  \{ D_1 \}], \ldots ,
                [(N - \Cs)  \cup  \{ D_k \}]\}$\\
                & where \Csp $\: = \{ D_1 \vee \ldots \vee D_k \}$\\
    \add: &  \Nodes $=  \{ N \cup \Csp \}$\\
    \delete: &  \Nodes $= \{ N  - \Csp \}$\\
    \marc: &  \Nodes $= \{ \lambda: N \}$
\end{tabular}

\medskip
\noindent The \marc{} action does not change the elements in $N$ but
it marks the node $N$ with the label $\lambda$.\footnote{As we will
see later, $\lambda$ can only be the label \emph{undefined}. When
clear from the context, we will represent a CIFF node omitting its
label.} All the actions, apart from the \replaceall{} action,
replace $N$ by a single successor node.

\noindent In the \replaceall{} action, \Csp{} consists of a single
conjunct in disjunctive form, i.e. $\Csp = \{ D_1 \vee \ldots \vee
D_k \}$. This action adds to $F$ a set \Nodes{} of $k$ successor
nodes, each of them obtained from $N$ by deleting \Cs{} and by
adding a single disjunct $D_i$.

\medskip
\noindent We are now ready to specify the proof rules in detail.

\medskip
\noindent In the presentation we are going to write
$\vec{t}=\vec{s}$ as a shorthand for $t_1=s_1\wedge\cdots\wedge
t_k=s_k$ (with the implicit assumption that the two vectors have the
same length), and $[\vec{X}\!/\vec{t}]$ for the substitution
$[X_1\!/t_1,\ldots,X_k\!/t_k]$. Note that $X$ and $Y$ will always
represent variables.

\medskip
\noindent Furthermore, in our presentation of the proof rules, we
abstract away from the order of conjuncts in the body of an
implication by writing the body of implications with the \apici{critical} conjunct
in the first position.


\medskip
\noindent Recall that, in writing the proof rules, we use
implicit variable quantification described in Definition
\ref{def:node}. 

\medskip
\noindent The first proof rule replaces an atomic conjunct in a node
$N$ by its iff-definition:

\ruledef{1}{Unfolding atoms}%
{$p(\vec{t})$}%
{$[p(\vec{X})\equivalent D_1\vee\cdots\vee D_n]\in\Th$}%
{\replace{} \{ $(D_1\vee\cdots\vee D_n)[\vec{X}\!/\vec{t}]$ \}}

\noindent Note that any variable in $D_1\vee\cdots\vee D_n$ is
implicitly existentially quantified in the resulting formula $F'$.

\noindent We assume that variable renaming may be applied so that
all existential variables have distinct names in the resulting CIFF
node.

\medskip
\noindent \emph{Unfolding} can be applied also to atoms occurring
in the body of an implication yielding one new implication for every
disjunct in the corresponding iff-definition:

\ruledef{2}{Unfolding within implications}%
{$(p(\vec{t})\wedge B)\imp H$}%
{$[p(\vec{X})\equivalent D_1\vee\cdots\vee D_n]\in\Th$}%
{\replace{} \{ $ [(D_1[\vec{X}\!/\vec{t}]\wedge B)\imp H], \ldots,
[(D_n[\vec{X}\!/\vec{t}]\wedge B)\imp H]$ \}}

\noindent Observe that, within $F'$, any variable in any $D_i$
becomes universally quantified with scope the implication in which
it occurs. Also in rule \textbf{R2} \emph{renaming} of variables is assumed, as
discussed for \textbf{R1}.

\medskip
\noindent The next rule is the \emph{propagation} rule, which allows
us to resolve an atom in the body of an implication in $N$ with a
matching atomic conjunct also in $N$.

\ruledef{3}{Propagation}%
{$[(p(\vec{t})\wedge B)\imp H],\quad p(\vec{s})$}%
{}
{\add{} \{ $(\vec{t}=\vec{s} \wedge B) \imp H$ \}}

\noindent Note that if $p$ has no arguments, $(\vec{t}=\vec{s}
\wedge B) \imp H$ should be read as $(\top \wedge B) \imp H$.

\medskip
\noindent The \emph{splitting} rule is the only rule performing a
\textbf{replace\_all} action. Roughly speaking it distributes a
disjunction over a conjunction.

\ruledef{4}{Splitting}%
{$D_1\vee\cdots\vee D_n$}%
{}%
{\replaceall{} \{ $D_1 \vee \ldots \vee D_n$ \}}

\medskip
\noindent The following \emph{factoring} rule can be used to
generate two cases, one in which the given abducible atoms unify and
one in which they do not:

\ruledef{5}{Factoring}%
{$p(\vec{t}),\quad p(\vec{s})$}%
{$p$ abducible}
{\replace{} \{ $[p(\vec{t})\wedge p(\vec{s})\wedge
(\vec{t}=\vec{s}\imp\bot)]\vee [p(\vec{t})\wedge\vec{t}=\vec{s}]$ \}
}

\medskip
\noindent The next set of CIFF proof rules are the \emph{constraint}
rules. They manage constraint atoms and they are, in a sense, the
interface to the constraint solver. They also deal with equalities
and CIFF disequalities (see Definition \ref{def:ciffdiseq}) which
can be delegated to the constraint solver if their arguments are in
the constraint domain $D(\Re)$. The formal definition of the proof
rules is quite complex, hence we first introduce some useful
definitions.

\begin{definition}[Basic c-atom]
A \emph{basic c-atom} is either a constraint atom, or an
equality atom of the form $A = B$ where $A$ and $B$ are not both variables,
and each is either a variable or a term ranging over
the chosen constraint domain $D(\Re)$.
\end{definition}

\noindent As an example, $X > 3$ and $X = 2$ are both basic
c-atoms, whereas $X = Y$ and $X = a$ are not (where $a \not\in
D(\Re)$).

\begin{definition}[basic c-conjunct and constraint variable]
A \emph{basic c-conjunct} is a basic c-atom which occurs as a CIFF
conjunct in a node.

\noindent A \emph{constraint variable} is a variable occurring in a
basic c-conjunct. 
\end{definition}

\noindent Note that a constraint variable is always an existentially
quantified variable with its scope the entire CIFF node in which it
occurs. This is because it must appear in a basic c-conjunct (i.e.
outside an implication).

\begin{definition}[c-atom and c-conjunct]
\label{def:c-atom}
A \emph{c-atom} is either a basic c-atom or a non-ground equality atom
of the form $A=B$ such that all the variables occurring in it are
constraint variables.

\noindent A \emph{c-conjunct} is a c-atom which occurs as a CIFF
conjunct in a node. 
\end{definition}

\medskip
\noindent We are now ready to present the first \emph{constraint}
proof rule.

\ruledef{6}{Case analysis for constraints}%
{$({\it Con}\wedge A) \imp B$}%
{${\it Con}$ is a c-atom}%
{\replace{} \{ $[{\it Con'}\wedge (A\imp B)]\vee \overline{{\it
Con'}}$ \}}

\noindent
where $Con'$ is $A \eqq B$ if $Con$ is $A=B$, and $Con'$ is $Con$ otherwise.\\
Observe that as $Con$ is a c-atom, all the variables
occurring in it are constraint variables, thus they are
existentially quantified.

\medskip
\noindent The next rule provides the actual \emph{constraint
solving} step itself. It may be applied to any set of c-conjuncts in
a node, but to guarantee soundness, eventually, it has to be applied
to the set of \emph{all} c-conjuncts in a node. To simplify
presentation, we assume that the constraint solver will fail
whenever it is presented with an ill-defined constraint such as,
say, ${\it bob}\leq 5$ (in the case of a numerical solver). For
inputs that are ``well-typed'', however, such a situation never
arises.

\ruledeflong{7}{Constraint solving}%
{${\it Con}_1, \ldots, {\it Con}_n$}%
{each $Con_i$ is a c-conjunct;} {$\{{\it Con'}_1,\ldots,{\it Con'}_n\}$ is not $\Re$-satisfiable}%
{\replace{} \{ $\bot$ \}}

\medskip
\noindent As in the case of the previous rule, ${\it Con'}_i$ is obtained from ${\it Con}_i$
by replacing all occurrences of $=$ with $\eqq$.

\medskip
\noindent The next proof rules deal with equalities (which are not
constraint atoms to be handled by the constraint solver) and they
rely upon the following rewrite rules which essentially implement
the term reduction part of the unification algorithm of
\cite{unification}:

\begin{enumerate}
\item[(1)] Replace $f(t_1,\ldots,t_k)=f(s_1,\ldots,s_k)$ by $t_1=s_1\wedge\cdots\wedge t_k=s_k$.
\item[(2)] Replace $f(t_1,\ldots,t_k)=g(s_1,\ldots,s_l)$ by $\bot$ if $f$ and $g$ are
distinct or $k\not= l$.
\item[(3)] Replace $t=t$ by $\top$.
\item[(4)] Replace $X=t$ by $\bot$ if $t$ contains $X$.
\item[(5)] Replace $t=X$ by $X=t$ if $X$ is a variable and $t$ is not.
\item[(6)] Replace $Y=X$ by $X=Y$ if $X$ is a \univquant\
variable and $Y$ is not.
\end{enumerate}

\noindent In the following \emph{equality rewriting} rules, we
denote as \RewEqs{e} the result of applying the above rewrite rules
(1)-(6) to the equality $e$. If no rewrite rule can be applied then
\RewEqs{e} $= e$.

\ruledef{8}{Equality rewriting in atoms}%
{$t_1=t_2$}%
{}
{\replace{} \{ \RewEqs{t_1=t_2} \}}

\ruledef{9}{Equality rewriting in implications}%
{($t_1=t_2\wedge B) \imp H$}%
{}
{\replace{} \{ (\RewEqs{t_1=t_2} $\wedge B) \imp H$ \}}

\medskip
\noindent The following two \emph{substitution rules} propagate
equalities to the rest of the node. In the first case we assume that
$N = (X = t \wedge Rest)$.

\ruledef{10}{Substitution in atoms}%
{$X=t, \quad  Rest$}%
{$X\not\in t$; $t$ is a Herbrand term}
{\replace{} \{ $X=t, \quad (Rest [X\!/t])$ \}}

\ruledef{11}{Substitution in implications}%
{$(X=t\wedge B)\imp H$}%
{$X$ \univquant; $X\not\in t$; $t$ is a Herbrand term} {\replace{}
\{ $(B\imp H)[X/t]$ \}}

\noindent Note that if $B$ is empty then $(B\imp H)[X/t]$ should be
read as $(\top \imp H)[X/t]$.

\medskip
\noindent If none of the \emph{equality rewriting} or
\emph{substitution} rules are applicable, then an equality in the
body of an implication may give rise to a \emph{case analysis:}

\ruledeflong{12}{Case analysis for equalities}%
{$(X=t\wedge B)\imp H$}%
{$(X=t\wedge B)\imp H$ is not of the form $X=t\imp\bot$; $X\not\in t$;}
{  $X$ is \existquant; $X=t$ is not a c-atom; \\ & \ \
$t$ is not a \univquant\ variable; \\ & \ \ $t$ is a Herbrand term}%
{\replace{}  \{ $[X=t \wedge (B \imp H)] \vee [X=t\imp\bot]$ \}}

\noindent Note that the variables which occur in $t$ become
existentially quantified in the first disjunct while in the second
disjunct each variable in $t$ maintains its original quantification.

\noindent The first condition of the rule avoids applying \emph{case
analysis} if the implication $(X=t\wedge B)\imp H$ is of the form
$X=t\imp\bot$. This is because, if it were applied, the resulting
first disjunct would become $[X=t \wedge (\top \imp \bot)]$ which is
trivially \emph{false}, while the second disjunct would become
$X=t\imp\bot$ itself. The other conditions guarantee that none of
the earlier rules are applicable.

\medskip
\noindent The next rule moves negative literals in the body of an
implication to the head of that implication:

\ruledef{13}{Negation rewriting}%
{$((A\imp\bot)\wedge B)\imp H$}%
{}%
{\replace{} \{ $B\imp (A\vee H)$ \}}

\noindent Note that if $B$ is empty then $B\imp (A\vee H)$ should
be read as $\top \imp (A\vee H)$.

\medskip
\noindent The following are \emph{logical simplification} rules.
\ruledef{14}{Logical simplification \#1}%
{ $\top$}%
{}%
{\delete{} \{ $\top$ \}}

\ruledef{15}{Logical simplification \#2}%
{$(\top\wedge B)\imp H$}%
{$B$ is not empty}%
{\replace{} \{ $B\imp H$ \}}

\ruledef{16}{Logical simplification \#3}%
{ $\bot\imp H$}%
{}%
{\delete{} \{ $\bot\imp H$ \}}

\ruledef{17}{Logical simplification \#4}%
{ $\top \rightarrow H$}%
{$H$ does not contain any universally quantified variable}%
{\replace{} \{ $H$ \}}

\noindent Note that the last simplification rule replaces an
implication with an empty body with its head as a CIFF conjunct.
This is done only if no universally quantified variables occur in
the head, otherwise we would have some universally quantified
variables outside implications in a node. For example, suppose we
applied the rule on $\top \rightarrow a(f(Y))$ where $Y$ is
universally quantified and $a$ is abducible. We would obtain
$a(f(Y))$ as a conjunct in a node, thus leading to two main
problems: (1) the variable quantification cannot be implicit and,
even worse, (2) the semantics should be extended to the case of
\emph{infinitely} many instantiations of abducible atoms in an
abductive answer.

\noindent The case where $H$ does have a universally quantified
variable is dealt with by the \emph{Dynamic Allowedness} rule, which
is used to identify nodes with problematic quantification patterns,
which could lead to floundering:

\ruledeflong{18}{Dynamic allowedness (DA)}%
{$B\imp H$}%
{either $B = \top$ or $B$ consists of constraint atoms alone;}{\ no other rule applies to the implication}%
{\marc{} \Undef}

\medskip
\noindent Due to the definition of the other CIFF proof rules, the
implication $B \imp H$ to which \textbf{DA} is applied to falls in
 one of the following cases:

\begin{enumerate}
  \item $B = \top$ and there is a universally quantified variable
  in $H$;
  \item there is a constraint atom in $B$ with an
universally quantified variable occurring in it.
\end{enumerate}

\noindent \textbf{DA} allows us to avoid obtaining infinitely many
abducible atoms in an abductive answer. For example, let us consider
an implication of the form $X > Y \imp H$ such that $X$ is
universally quantified. Depending on $D(\Re)$, there could be
infinitely many instances of $X$ satisfying the c-atom and CIFF
should handle \emph{all} those cases. However, we believe that
\textbf{DA} could be relaxed, in particular for those implications
falling in case 2 above. Consider, for example, the following
implication:

\[ X > 3 \wedge X < 100 \rightarrow a(X) \]

\noindent where $X$ is universally quantified and $a$ is an
abducible predicate. If $D(\Re)$ is the set of all integers, there
is a finite set of abducible atoms satisfying the implication, i.e.
the set $\{ a(4), a(5), \ldots, a(99) \}$. However, \textbf{DA}
marks a node with this implication as undefined due to the presence
of $X$. The relaxation of \textbf{DA} is not in the scope of this
paper.

\medskip
\noindent The CIFF proof rules are summarized in Table
\ref{CIFFproofrules} where the rules drawn from the IFF procedure
are indicated by \apici{IFF} on the right-hand side. It is worth
noting that the four \textbf{Logical Simplification} rules are a
reformulation of the corresponding IFF rules where, in particular,
\textbf{Logical Simplification \#4} checks for the quantification of
the variables in the head of an implication for managing correctly
the floundering problem. Moreover, \textbf{Case analysis for
equalities} is a slight extension of the corresponding IFF rule for
handling c-atoms.

\begin{table}
\caption{CIFF proof rules}\label{CIFFproofrules}
\begin{tabular}{llr}
\hline\hline
 \textbf{R1} & Unfolding atoms & IFF\\
 \textbf{R2} & Unfolding in implications & IFF\\
 \textbf{R3} & Propagation & IFF\\
 \textbf{R4} & Splitting & IFF\\
 \textbf{R5} & Factoring & IFF\\
 \textbf{R6} & Case analysis for constraints \\
 \textbf{R7} & Constraint solving \\
 \textbf{R8} & Equality rewriting in atoms & IFF\\
 \textbf{R9} & Equality rewriting in implications & IFF\\
 \textbf{R10} & Substitution in atoms & IFF\\
 \textbf{R11} & Substitution in implications & IFF\\
 \textbf{R12} & Case analysis for equalities & IFF\\
 \textbf{R13} & Negation rewriting & IFF\\
 \textbf{R14} & Logical Simplification \#1 & IFF\\
 \textbf{R15} & Logical Simplification \#2 & IFF\\
 \textbf{R16} & Logical Simplification \#3 & IFF\\
 \textbf{R17} & Logical Simplification \#4 & IFF\\
 \textbf{R18} & Dynamic Allowedness\\
\hline\hline
\end{tabular}

\end{table}

\subsection{CIFF Derivation and Answer Extraction}\label{sec:ciffder}
\noindent The CIFF proof rules are the building blocks of a
\emph{CIFF derivation} which defines the process of computing
answers with respect to a framework \CIFFprog{} and a query $Q$.

\noindent Prior to defining a CIFF derivation formally, we introduce
some useful definitions.

\begin{definition}[Failure and undefined CIFF nodes] A CIFF node $N$ which contains $\bot$ as an atomic CIFF conjunct is called
a \emph{failure CIFF node}. A CIFF node $N$ marked as \Undef{} is
called an \emph{undefined CIFF node}. 
\end{definition}

\begin{definition}[CIFF selection function]\label{selection}
Let $F$ be a CIFF formula. We define a \emph{CIFF selection
function} \Sel{} as a function such that:

\begin{center} \SelA{F} $ = \langle N, \phi, \Cs \rangle $\end{center}

\noindent where $N$ is a CIFF node in $F$, $\phi$ is a CIFF proof
rule and \Cs{} is a set of CIFF conjuncts in $N$ such that \Cs{} is
a \emph{rule input} for $\phi$.

\end{definition}

\noindent In the sequel we assume that selection functions, given a
CIFF formula $F$, always select a triple $\langle N, \phi, \Cs
\rangle$ whenever a rule is applicable to $F$.


\medskip
\noindent We are now ready to define a \emph{CIFF
pre-derivation} and a \emph{CIFF branch}.

\begin{definition}[CIFF Pre-derivation and initial formula]\label{prederivation} Let \CIFFprog{} be a CIFF framework, let $Q$ be a query
and let \Sel{} be a \emph{CIFF selection function}. A \emph{CIFF
pre-derivation} for $Q$ with respect to \CIFFprog{} and \Sel{} is a
(finite or infinite) sequence of CIFF formulae $F_1, F_2, \ldots,
F_i, F_{i+1} \ldots$ such that each $F_{i+1}$ is obtained from $F_i$
through \Sel{} as follows:

\begin{itemize}
  \item $F_1 = \{ N_1 \} = \{ Q \cup IC \}$, where $Q$ and $IC$ are
  treated as sets of CIFF conjuncts, (we will refer to $F_1$ as the
  \emph{initial formula} of a CIFF pre-derivation)
  \item $\Sel$$(F_i) = \langle N_i, \phi_i, \Cs_i \rangle$ such that $N_i$
is neither an undefined CIFF node nor a failure CIFF node and
\medskip
\item \ruleappder
\end{itemize}

%

\end{definition}

\noindent The construction of a pre-derivation can be interpreted as
the construction of an or-tree rooted at $N_1$ and whose nodes are
CIFF nodes. Roughly speaking, the whole or-tree can be seen as a
search tree for answers to the query. Note that all the variables in
the query are existentially quantified in $N_1$ because the
allowedness conditions of Definition \ref{def:CIFFallowed} impose
that each variable in $Q$ occurs in an atomic conjunct of $Q$.

\noindent CIFF formulas $F_i$ in a pre-derivation correspond to
successive frontiers of the search tree. Each derivation step is
done by applying (through \Sel{}) the \emph{selected} proof rule  on
a set $\Cs$ of CIFF conjuncts within a node $N$ in a frontier. The
resulting frontier is obtained by replacing $N$ by the set of
\emph{successor nodes} \Nodes{}.

\begin{definition}[Successor Nodes in a CIFF pre-derivation]
\noindent Let \Der{} be a CIFF pre-derivation for a query $Q$ with
respect to a CIFF framework \CIFFprog{} and a selection function
\Sel.

\noindent We say that \Nodes{} is the set of \emph{successor nodes
of $N$ in \Der}, iff

\begin{itemize}
  \item $\Sel$$(F_i) = \langle N, \phi_i, \Cs_i \rangle$,

  \medskip
  \item \ruleappargs{F_i}{$N,\Cs_i$}{$\phi_i$}{F_{i+1}}, and

\medskip
 \item for each $N' \in F_{i+1}$ such that $N'
\not\in
  F_{i} \backslash \{ N \}$, then $N' \in \Nodes{}$.
\end{itemize}

\noindent Moreover we say that a node $N'$ in \Nodes{} is a
\emph{successor node of $N$ in \Der}.
\end{definition}

%
%
%

\begin{definition}[CIFF branch] Given a CIFF pre-derivation \Der{} $= F_1, F_2, \ldots, F_i, F_{i+1}\ldots$,
a \emph{CIFF branch} \Branch{} in \Der{} is a (finite or infinite)
sequence of CIFF nodes $N_1, N_2, \ldots, N_i, N_{i+1}\ldots$ such
that each $N_i \in F_i$ and each $N_{i+1}$
is a CIFF successor node of $N_i$ in \Der. 
\end{definition}

\noindent The next step, finally, is the definition of a CIFF
derivation.

\begin{definition}[CIFF derivation]\label{derivation}
Let \CIFFprog{} be a CIFF framework, let $Q$ be a query and let
\Sel{} be a \emph{CIFF selection function}. A \emph{CIFF derivation}
\Der{} for $Q$ with respect to \CIFFprog{} and \Sel{} is a CIFF pre-derivation
$F_1, F_2, \ldots$ such that for each CIFF branch \Branch{} in
\Der{} if

\begin{itemize}
  \item $\Sel(F_i) = \langle N_i, \phi, \Cs \rangle$,
  \item $\Sel(F_j) = \langle N_j, \phi, \Cs \rangle$,
  \item $N_i \in {\cal B}$,
  \item $N_j \in {\cal B}$ and
  \item $i \neq j$
\end{itemize}

\noindent then $\phi \not\in $ \{\textbf{Propagation},
\textbf{Factoring}, \textbf{Equality rewriting in atoms},
\textbf{Equality rewriting in implications}, \textbf{Substitution in
atoms}\}. 
\end{definition}

%

\noindent Informally, a derivation is a pre-derivation such that in
each branch certain proof rules can be applied only once to a given
set of selected CIFF conjuncts. This is because those rules can
produce loops if they are applied repeatedly to the same set of
conjuncts\footnote{Note, however, that they could be applied to
different \emph{copies} of a set of conjuncts.}. The concept of
successor nodes in a pre-derivation is valid also for a derivation. Where
it has no impact, we will omit the selection function when we refer
to a derivation.

\begin{example}
Consider the following framework \abdciffframe{}:

\[\begin{array}{l@{\quad}l}
\Th: & p \leftrightarrow \top\\[1pt]

\Abd: & \{ a\}\\[1pt]

\IC: & p \;\imp\; a\\[1pt] \end{array}\]

\noindent The following is a pre-derivation \Der{} for the query $Q
= p$.


\[\begin{array}{l@{\;\;}l@{\;\;}l}
F_1 = & \{\{ p, [p \imp a] \}\} & \mbox{[\textbf{Init}]} \\
F_2 = & \{\{ p, [p \imp a], [\top \imp a] \}\} & \mbox{[\textbf{R3}]} \\
F_3 = & \{\{ p, [p \imp a], [\top \imp a], [\top \imp a] \}\}& \mbox{[\textbf{R3}]} \\
& \vdots & \\
\end{array}\]

\noindent The \textbf{Propagation} rule \textbf{R3} can be applied
repeatedly to the integrity constraint giving rise to an infinite
pre-derivation which should be avoided in a derivation\footnote{The
example shows the need of \emph{multisets} for representing
correctly CIFF formulae and CIFF nodes.}.
\end{example}

\begin{definition}[Successor CIFF Derivation]
Let \Der $= F_1, \ldots, F_i$ be a CIFF derivation, let \Sel{} be a
CIFF selection function and let $N \in F_i$. We say that $\Der' =
F_1, \ldots, F_{i+1}$ is a \emph{successor CIFF derivation via $N$
of \Der{}} iff

\begin{itemize}
  \item $\Sel$$(F_i) = \langle N, \phi_i, \Cs_i \rangle$,
  \medskip
  \item \ruleappargs{F_i}{$N,\Cs_i$}{$\phi_i$}{F_{i+1}}, and
  \medskip
  \item $\Der'$ is a CIFF derivation.
\end{itemize}

\end{definition}

\begin{definition}[Leaf and successful CIFF nodes]
Let \Der $= F_1, \ldots, F_i$ be a CIFF derivation. A CIFF node $N$
in $F_i$ is a \emph{leaf CIFF node} iff

\begin{itemize}
    \item it is a \emph{failure CIFF node} or
    \item it is an \emph{undefined CIFF node} or
    \item there exists no successor CIFF derivation via $N$ of \Der.
\end{itemize}

\noindent A \emph{leaf node} which is neither a failure CIFF node
nor an undefined CIFF node is called a \emph{successful CIFF node}.
\end{definition}

\noindent We are now ready to introduce the following
classifications of \emph{CIFF branches} and \emph{CIFF derivations}.

\begin{definition}[Failure, undefined and successful CIFF branches]
Let \Der{} be a CIFF derivation and let $\Branch = \sequenza{N}{k}$
 be a CIFF branch in \Der{}. We say that \Branch{} is

\begin{itemize}
    \item a \emph{successful CIFF branch} if $N_k$ is a successful CIFF node;
    \item a \emph{failure CIFF branch} if $N_k$ is a failure CIFF node;
    \item an \emph{undefined CIFF branch} if $N_k$ is an undefined CIFF
    node.
\end{itemize}

\end{definition}

\begin{definition}[Failure and Successful CIFF Derivations]
Let \Der{} be a CIFF derivation. \Der{} is called a \emph{successful
CIFF derivation} iff it contains at least one \emph{successful CIFF
branch}. \Der{} is called a \emph{failure CIFF derivation} iff all
its branches are \emph{failure CIFF branches}. 
\end{definition}

\noindent Intuitively, an abductive answer to a query $Q$ can be
extracted from a successful node of a \emph{successful derivation}.
Formally:

\begin{definition}[CIFF Extracted Answer]\label{def:extractedanswer}
Let \abdciffframe{} be a CIFF framework and let $Q$ be a CIFF query.
Let \Der{} be a successful CIFF derivation for $Q$ with respect to
\abdciffframe. A \emph{CIFF extracted answer} from a successful node
$N$ of \Der{} is a pair

\[ \answerciff \]

\noindent where $\Delta$ is the set of abducible atomic conjuncts in
$N$, and $C = \langle \Gamma, E, DE \rangle$ where:

\begin{itemize}
\item $\Gamma$ is the set of all the c-conjuncts in $N$,
\item $E$ is the set of all the \emph{equality atoms} (i.e. equalities over Herbrand terms) in $N$,
\item $DE$ is the set of all the CIFF
disequalities in $N$.
\end{itemize}

\end{definition}

\noindent The soundness of the CIFF proof procedure with respect to
the notion of $\Re$-satisfiability and the three-valued completion
semantics is the subject of the next section. The idea is to show
that CIFF extracted answers correspond to abductive answers
with constraints in the sense of Definition \ref{def:abdcanswer}. 

\begin{example}
Consider the following framework \abdciffframe, obtained from the
abductive logic program with constraints of Example
\ref{ex:example1}, and the following query $Q$:

\[\begin{array}{l@{\quad}l}
 \Th: & p(T) \leftrightarrow T = X \wedge q(T_1,T_2) \wedge T_1\!\!<\!X \wedge X\!\!<\!8 \\
      & q(X,Y) \leftrightarrow X = X_1 \wedge Y = X_2 \wedge s(X_1,a) \\[1pt]
 \Abd: & \{ r, s\}\\[1pt]
 \IC: & r(Z) \;\imp\; p(Z)\\[1pt]
 Q: & r(Y)
\end{array}\]

\noindent The following is a CIFF derivation \Der{} for $Q$ with
respect to \abdciffframe{}:


\begin{small}
\[\begin{array}{l@{\;\;}l@{\;\;}r}
F_1 = & \{\{ r(Y), [r(Z) \imp p(Z)] \}\} & \mbox {[\textbf{Init}]} \\
F_2 = & \{\{ r(Y), [Z=Y \imp p(Z)], [r(Z) \imp p(Z)] \}\} & \mbox {[\textbf{R3}]} \\
F_3 = & \{\{ r(Y), [\top \imp p(Y)], [r(Z) \imp p(Z)] \}\} & \mbox {[\textbf{R11}]} \\
F_4 = & \{\{ r(Y), p(Y), [r(Z) \imp p(Z)] \}\} & \mbox {[\textbf{R17}]} \\
F_5 = & \{\{ r(Y), Y = X, q(T_1,T_2), T_1\!\!<\!X, X\!<\!8, [r(Z) \imp p(Z)] \}\} & \mbox {[\textbf{R1}]} \\
F_6 = & \{\{ r(X), Y = X, q(T_1,T_2), T_1\!\!<\!X, X\!<\!8, [r(Z) \imp p(Z)] \}\} & \mbox {[\textbf{R10}]} \\
F_7 = & \{\{ r(X), Y = X, T_1 = V, T_2 = W, s(V,a), T_1\!\!<\!X, X\!<\!8, [r(Z) \imp p(Z)] \}\} & \mbox {[\textbf{R1}]} \\
F_8 = & \{\{ r(X), Y = X, T_1 = V, T_2 = W, s(V,a), V\!\!<\!X,
X\!<\!8, [r(Z) \imp p(Z)] \}\} & \mbox {[\textbf{R10}]}


\end{array}\]
\end{small}

\noindent No more new rules can be applied to the only node in $F_8$
and this is neither a failure node nor an undefined node. Hence, it
is a \emph{successful node} from which we extract the following
answer:

\[ \langle \{r(X),s(V,a)\}, C \rangle \]

\noindent where $C = \langle \Gamma, E, DE \rangle$ is:

\[\begin{array}{l@{\;\;}l}
\Gamma: & \{Y = X, T_1 = V, V\!\!<\!X, X\!<\!8\} \\
E: & \{T_2 = W\} \\
DE: & \oslash \\
\end{array}\]

\noindent Indeed, note that the \emph{abductive answers with constraints} given
in Example~\ref{ex:example1} are instances of the above extracted answer. 
\end{example}

\begin{example}
Consider the following framework \abdciffframe{} (where we assume a
constraint structure $\Re$ over integers with the usual relations
and functions), and the following query $Q$:

\[\begin{array}{l@{\quad}l}
 \Th: & p(X) \leftrightarrow X = Z \wedge a(Z) \wedge Z < 5 \\
 \Abd: & \{ a\}\\[1pt]
 \IC: & a(2) \;\imp\; \bot\\[1pt]
 Q: & p(Y)
\end{array}\]

\noindent The following is a CIFF derivation \Der{} for $Q$ with
respect to \abdciffframe:

\begin{small}
\[\begin{array}{l@{\;\;}l@{\;\;}r}
F_1 = & \{\{ p(Y), [a(2) \imp \bot] \}\} & \mbox {[\textbf{Init}]} \\
F_2 = & \{\{ Y = Z, a(Z), Z < 5, [a(2) \imp \bot] \}\} & \mbox {[\textbf{R1}]} \\
F_3 = & \{\{ Y = Z, a(Z), Z < 5, [a(2) \imp \bot], [2 = Z \imp \bot] \}\} & \mbox {[\textbf{R3}]} \\
F_4 = & \{\{ Y = Z, a(Z), Z < 5, [a(2) \imp \bot], [Z = 2 \imp \bot] \}\} & \mbox {[\textbf{R9}]} \\
F_5 = & \{\{ Y = Z, a(Z), Z < 5, [a(2) \imp \bot], [Z \neq 2 \vee [Z \eqq 2, (\top \imp \bot)]] \}\} & \mbox {[\textbf{R6}]} \\
F_6 = & \{ \{ Y = Z, a(Z), Z < 5, Z \neq 2, [a(2) \imp \bot] \},\\
& \ \: \{ Y = Z, a(Z), Z < 5, Z \eqq 2, [a(2) \imp \bot], (\top \imp \bot) \}\} & \mbox {[\textbf{R4}]} \\
F_7 = & \{ \{ Y = Z, a(Z), Z < 5, Z \neq 2, [a(2) \imp \bot] \},\\
& \ \: \{ Y = Z, a(Z), Z < 5, Z \eqq 2, [a(2) \imp \bot], \bot \}\} & \mbox {[\textbf{R17}]} \\
\end{array}\]
\end{small}

\noindent Note that only the \textbf{Case analysis for constraints}
rule (\textbf{R6}) can be applied to $F_4$ because the variable $Z$
is a constraint variable. Hence $Z = 2$ is a c-atom (see Definition \ref{def:c-atom}) and thus the
\textbf{Case analysis for equalities} rule (\textbf{R12}) cannot be
applied to $F_4$.

\noindent No more rules can be applied to both nodes in $F_7$. The
first node is neither a failure node nor an undefined node. Hence,
it is a \emph{successful node} from which we extract the following
answer:

\[ \langle \{a(Z)\}, \langle \{Y = Z, Z < 5, Z \neq 2\}, \oslash, \oslash \rangle \rangle \]

%

\end{example}


\section{Correctness of the CIFF Proof Procedure}\label{sec:soundness}
\noindent As anticipated in the previous section, the CIFF proof
procedure is sound with respect to the three-valued completion
semantics, i.e. each CIFF extracted answer is indeed a CIFF correct
answer in the sense of definition \ref{def:abdcanswer}. All the
results stated in this section (and whose proofs are given in
\ref{AppendixA}) are based upon the results given in
\cite{iffthesis} for the IFF proof procedure.

\begin{theorem}[CIFF Soundness]\label{theo:soundness}
Let \AbdCprog{} be an abductive logic program with constraints such
that the corresponding CIFF framework is \abdciffframe{}. Let
\answerciff, where $C = \langle \Gamma, E, DE \rangle$, be a CIFF
extracted answer from a successful CIFF node in a CIFF derivation
with respect to \abdciffframe{} and a CIFF query $Q$. Then there
exists a ground substitution $\sigma$ such that
 \AnswerC{} is an abductive answer with constraints to $Q$ with respect to \AbdCprog.

\end{theorem}

\medskip
\noindent The proof of the theorem relies upon the following
propositions.
\noindent The first proposition shows that given a CIFF extracted
answer \answerciff{} there exists a substitution satisfying
all the constraint atoms, equality atoms and CIFF disequalities in $C$. 
%
%

\begin{proposition}\label{lemmasub1}
Let \answerciff{} be a CIFF extracted answer from a successful CIFF
node $N$, where $C = \answerC$. Then:

\begin{enumerate}
  \item there exists a ground substitution $\theta$ such that $\theta \models_{3(\Re)} \Gamma$, and

  \item for each such ground substitution $\theta$, there exists a
  ground substitution $\sigma$ such that
%
  \[ \theta\sigma \models_{3(\Re)} \Gamma \cup E \cup DE \]
\end{enumerate}

%
%

\end{proposition}

\begin{example}
Given $\Gamma = \{ 2 \leq T, T < 4\}$, $E = \{ X = f(Y), Z = g(V)\}$
and $DE = \{ (Y = h(W,V)) \rightarrow \bot\}$, we have that both
$\theta_1 = \{ T/2 \}$ and $\theta_2 = \{ T/3 \}$ satisfy $\Gamma$
and they contain all the possible assignments for $T$ (given that
$D(\Re)$ is the set of all integers). We can obtain a ground
substitution $\theta_1\sigma$ (with $\sigma = \sigma_{DE} \cup
\sigma_{E}$) as follows:

\begin{enumerate}
  \item $\sigma_{DE} = \{ Y/r(c) \}$ obtaining
        $S_1 = ((E \cup DE)\theta)\sigma_{DE} =$\\
        \medskip
        $\{ X = f(r(c)), Z = g(V), (r(c) = h(W,V)) \rightarrow \bot
        \}$
        \medskip
  \item the second step is to assign the corresponding terms to $X$
  and $Z$ obtaining
        \[ S_2 = \{ f(r(c)) = f(r(c)), g(V) = g(V), (r(c) = h(W,V)) \rightarrow \bot \} \]
  \item finally we assign new terms with fresh functions to the remaining existentially quantified variable $V$, e.g.
  $\sigma_{E} = \{ V/t(c) \}$ obtaining
        \[ S_3 = \{ f(r(c)) = f(r(c)), g(t(c)) = g(t(c)), (r(c) = h(W,t(c))) \rightarrow \bot \} \]
\end{enumerate}

\noindent The set $S_3$ is clearly entailed by CET. Note that we do
not care about the universally quantified variable $W$ in $S_3$.
This is because

\[(r(c) = h(W,t(c))) \rightarrow \bot\]

\noindent is entailed by CET for any assignment to $W$, due to the
fact that $r$ and $h$ are distinct function symbols.

\medskip
\noindent Similarly, we can obtain another ground substitution using
$\theta_2$. 
\end{example}

\medskip
\noindent The next proposition directly extends the above result to
the set $\Delta$ of a CIFF extracted answer.

\begin{proposition}\label{lemmasub2}
Let \answerciff{} be a CIFF extracted answer from a successful CIFF
node $N$ where $C = \answerciffC$. For each ground substitution
$\sigma'$ such that $\sigma' \models_{3(\Re)} \answerciffCcup$,
there exists a ground substitution $\sigma$ which extends $\sigma'$
for the variables that are in $\Delta$ but not in $C$ such that

\begin{enumerate}
    \item $\sigma' \subseteq \sigma$
    \item $\Delta\sigma \models_{3(\Re)} \Delta \cup
    \answerciffCcup$.
\end{enumerate}

\end{proposition}

%
%
%

\medskip
\noindent The third proposition shows that the CIFF proof rules are
indeed equivalence preserving rules with respect to the three-valued
completion semantics. This a basic requirement to prove the
soundness of CIFF.

\begin{proposition}[Equivalence Preservation]\label{lemmaeq}
Given an abductive logic program with constraints \AbdCprog, a CIFF
node $N$ and a set of CIFF successor nodes \Nodes{} obtained by
applying a CIFF proof rule $\phi$ to $N$, it holds that:

\begin{center}
$P \models_{3(\Re)} N $ \quad iff \quad $P \models_{3(\Re)}
\Nodes^{\vee}$
\end{center}

\noindent where $\Nodes^{\vee}$ is the disjunction of the nodes in
\Nodes. 
\end{proposition}

\begin{corollary}[Equivalence Preservation of CIFF Formulae]\label{coroleq}
Let \AbdCprog{} be an abductive logic program with constraints, $F$
a CIFF formula and \Sel{} any CIFF selection function. Let $\Sel(F)
= \langle N, \phi, \Cs \rangle$ and $F'$ the result of applying
$\phi$ to $N$ in $F$. Then:

\begin{center}
$P \cup IC \models_{3(\Re)} F$ \quad iff \quad $P \cup IC
\models_{3(\Re)} F'$, i.e.
\end{center}

\[P \cup IC \models_{3(\Re)} (F \leftrightarrow F').\]

\end{corollary}


\noindent The CIFF soundness in Theorem \ref{theo:soundness}
concerns only those branches of a CIFF successful derivations whose
leaf node is a CIFF successful node. It implies that abductive
answers with constraints can be obtained also by those derivations
which contain failure and undefined branches but which have at least
a successful branch. We also prove the following notion of soundness
regarding failure CIFF derivations.

\begin{theorem}[Soundness of failure]\label{theo:soundnessfail}
Let \AbdCprog{} be an abductive logic program with constraints such
that the corresponding CIFF framework is \abdciffframe{}. Let \Der{}
be a failure CIFF derivation with respect to \abdciffframe{} and a
query $Q$. Then:

 \[ P  \cup IC \models_{3(\Re)} \neg Q. \]

\end{theorem}

\noindent Note that there is a class of CIFF derivations for which a
soundness result cannot be stated, i.e. all the derivations
containing only undefined and failure branches. The meaning of such
 CIFF derivations is that for each branch, no CIFF answer can be
extracted, but there are some branches (undefined branches) for
which neither failure nor success is ensured. The presence of an
undefined branch is due to the application of the \textbf{Dynamic
Allowedness} rule and, as we have seen at the end of Section
\ref{sec:ciffrules}, this could lead to infinite sets of abducibles
in the answers.

\medskip
\noindent Concerning completeness, CIFF inherits the completeness
results for IFF in \cite{iffthesis} for the class of allowed IFF
frameworks. In \cite{iffthesis}, the only requirement for ensuring
completeness is the use of a \emph{fair} selection function, i.e. a
selection function that ensures that any node to which a proof rule
can be applied is eventually selected
in each branch of a derivation. This condition is also
required in the case of CIFF. To illustrate \emph{fairness}, suppose
we have the following iff-definitions

\[\begin{array}{l}

q \leftrightarrow p \vee a \\
p \leftrightarrow p
\end{array}\]

\noindent where $a$ is an abducible predicate. Consider the query
$q$ and an empty set of integrity constraints. After the unfolding
of $q$, the IFF proof procedure would return the abductive answer
$a$ if the second disjunct is eventually selected, but it loops
forever in the other case. A \emph{fair} selection function ensures
that the second disjunct is eventually selected during a derivation.

\noindent For the class of IFF allowed frameworks, a CIFF derivation
is exactly an IFF derivation as there are no constraint atoms in the
framework. Moreover, the \textbf{Dynamic allowedness} rule can never
apply in a derivation due to the following lemma, stating that for
the of class CIFF statically allowed frameworks and queries
(see Definition
\ref{def:StaticAllowedness}) there does not exist
a CIFF derivation in which \textbf{Dynamic allowedness} is
applied.


\begin{lemma}[Static Allowedness lemma]\label{lemma:static_allowedness}
Let \AbdCprog{} be an abductive logic program with constraints such
that the corresponding CIFF framework \abdciffframe{} and the query
$Q$ are both CIFF statically allowed. Then, given any CIFF derivation
$F_1, F_2, \ldots$ with respect to \AbdCprog{} and $Q$, and any selection
function \Sel: it is never the case that
$\Sel(F_i) = \langle N_i, R18, \Cs \rangle$ for any $F_i$, where $R18$ is the
Dynamic allowedness rule.
\end{lemma}

\noindent Indeed, the above lemma trivially applies also to IFF
allowed frameworks.

\noindent As a consequence, we can state the following result.

\begin{theorem}[CIFF completeness for IFF allowed frameworks]\label{theo:IFFcompleteness}
Let \AbdCprog{} be an abductive logic program without constraints
such that the corresponding CIFF framework \abdciffframe{} and the
query $Q$ do not contain constraint atoms and they are IFF allowed.

\noindent If there exists an abductive answer with constraints
\ensuremath{\langle \Delta, \sigma, \oslash \rangle} for $Q$ with
respect to \AbdCprog, then there exists a CIFF derivation \Der{}
for $Q$ with respect to \abdciffframe{} and to a fair CIFF selection
function \Sel{} such that

\begin{itemize}
  \item \ensuremath{\langle \Delta', \langle \oslash,
E, DE \rangle \rangle}, can be extracted from a successful CIFF node
in \Der{}; and
  \item there exists a ground substitution $\sigma'' \supseteq \sigma$
such that
\begin{itemize}
 \item $\Prog \cup \Delta'\sigma'' \models Q\sigma''$
 \item $\Prog \cup \Delta'\sigma'' \models IC$
 \item $\Delta'\sigma'' \subseteq \Delta\sigma''$.
\end{itemize}
\end{itemize}
\end{theorem}

\noindent Considering the whole class of CIFF frameworks, we cannot
formulate a full completeness theorem for CIFF because, tackling the
allowedness problem dynamically, we could obtain undefined
derivations, even with a \emph{fair} selection function.

\begin{example}
Consider the following framework \abdciffframe{} where we assume an
arithmetical constraint over integers in which $>$ has the expected
meaning:

\[\begin{array}{l@{\quad}l}
\Prog: & p(Y) \leftarrow a(Y)\\[1pt]
\Th: & p(X) \leftrightarrow [X = Y \wedge a(Y)]\\[1pt]

\Abd: & \{ a\}\\[1pt]

\IC: & V > 2 \;\imp\; a(V)\\[1pt] \end{array}\]

\noindent The following is a CIFF derivation \Der{} for the empty
query.

\[\begin{array}{l@{\;\;}l@{\;\;}l}
F_1 = & \{\{ [V > 2 \;\imp\; a(V)] \}\} & \mbox{[\textbf{Init}]} \\
F_2 = & \{ undefined : \{ [V > 2 \;\imp\; a(V)] \}\} & \mbox{[\textbf{R18}]} \\
\end{array}\]

\noindent The only rule applicable to $F_1$ is
the \textbf{Dynamic allowedness} rule due to the presence of $V$ in
the constraint atom $V > 2$. Note that the existence of infinite
values for $V$ greater than $2$ would give rise to an infinite set
of abducibles arising from $a(V)$ in the head of the implication.
\end{example}

\noindent However, we can state a weak completeness theorem for the
CIFF proof procedure if we assume CIFF derivations without undefined
branches. The result is analogous to the completeness result shown
for the \Asys{} \cite{Asystemthesis,Asystem1}.

\begin{theorem}[Weak CIFF Completeness]\label{theo:completeness}
Let \AbdCprog{} be an abductive logic program with constraints with
the corresponding CIFF framework \abdciffframe{} and let $Q$ be a
CIFF query. Let \Der{} be a finite CIFF derivation with respect to
\abdciffframe{} and $Q$ such that each branch in \Der{} is either a failure
or a successful branch. Then:

\begin{enumerate}
\item if  $P  \cup IC \models_{3(\Re)} \neg Q$ then all the branches
of \Der{} are failure branches; and

\item if $P  \cup IC \not\models_{3(\Re)} \neg Q$ (i.e. $P  \cup IC \cup Q$ is satisfiable) then there exists a successful
branch in \Der.

\end{enumerate}
\end{theorem}

%

\noindent The above result gives rise to the following
completeness theorem for the CIFF proof procedure.

\begin{theorem}[Weak CIFF Completeness for CIFF statically allowed frameworks]\label{theo:static_completeness}
Let \AbdCprog{} be an abductive logic program with constraints such
that the corresponding CIFF framework \abdciffframe{} and the query
$Q$ are both CIFF statically allowed. Let \Der{} be a finite CIFF
derivation with respect to \abdciffframe{} and $Q$. Then:

\begin{enumerate}
\item if  $P  \cup IC \models_{3(\Re)} \neg Q$ then all the branches
of \Der{} are failure branches; and

\item if $P  \cup IC \not\models_{3(\Re)} \neg Q$ (i.e. $P  \cup IC \cup Q$ is satisfiable) then there exists a successful
branch in \Der.

\end{enumerate}
\end{theorem}

\noindent All the correctness results so far focus on the
three-valued completion semantics. However, it is worth noting
that both IFF and CIFF are sound with respect to the well-founded
semantics \cite{WellFound}, since the well-founded model is a
three-valued model of the completion of a logic program
\cite{WellFound}. However IFF (and thus CIFF for the class of IFF
allowed frameworks) is not complete with respect to that semantics.
Indeed, considering the iff-definition

\[\begin{array}{l}
p \leftrightarrow p
\end{array}\]

\noindent the negative literal $\neg p$ holds with respect to the
well-founded semantics while $p$ is \emph{undefined} with respect to
the three-valued completion semantics. Accordingly, both IFF and
CIFF fail to terminate for the query $\neg p$.


\section{The CIFF System}
\label{system}

The \CIFFSys{} is a SICStus
Prolog\footnote{\url{http://www.sics.se/isl/sicstuswww/site/index.html}}
implementation of CIFF. We rely upon the SICStus CLPFD solver
integrated in the platform. This is a very fast and reliable
constraint solver for finite domains \cite{SicstusCLPFD}. The
version of the system described here is version 4.0 whose engine has
been almost completely rewritten with respect to older versions
\cite{CIFFSystem,CIFFARW}, in order to improve efficiency. 

\medskip
\noindent Here we give a brief general description of the \CIFFSys.
Further details can be found in \cite{CIFFThesis} and in the CIFF
user manual \cite{CIFFManual}.


\medskip
\noindent The main predicate, to be run at Prolog top-level is

{\tt run\_ciff( +ALP, +Query, -Answer)}

\noindent where {\tt ALP} is a list of {\tt .alp} files containing
an abductive logic program with constraints\footnote{All the files
in the {\tt ALP} list together represent a single abductive logic
program with constraint. This is to facilitate writing CIFF
applications. A typical example is a list with two elements where
one {\tt .alp} file contains the clauses and the integrity
constraints which specify the problem and the other file contains
the specification of the particular problem instance. In this way
the first file could be reused for other
 instances.}, {\tt Query} is a CIFF query
and {\tt Answer} will be instantiated to either a CIFF extracted
answer (see Definition \ref{def:extractedanswer}) or to the special
atom {\tt undefined} if an allowedness condition is not met. A CIFF
extracted answer is represented by a triple, namely a list of
abducible atoms $\Delta$, a list of CIFF disequalities $DE$ and
finally a list of finite domain constraints $\Gamma$. The set of
equalities $E$ is not returned as the final substitution (in $E$) is
directly applied by the system. Further answers are returned via
Prolog backtracking. If no (further) answer is found, the system
fails,
returning the control to the Prolog top-level.

%
%
%
%

\medskip
\noindent Each abductive logic program with constraints (ALPC)
consists of the following components, which could be placed in any
position in any {\tt .alp} file:

\begin{itemize}
  \item Declarations of abducible predicates, using the predicate
  {\tt abducible}. For example an abducible predicate {\tt abd} with
  arity 2, is declared via

\medskip
  {\tt abducible(abd(\_,\_))}.
\medskip
  \item Clauses, represented as
  \begin{verbatim}
A :- L1, ..., Ln.
\end{verbatim}
  \item Integrity constraints, represented as
  \begin{verbatim}
[L1, ..., Lm] implies [A1, ..., An].
\end{verbatim}

\noindent where the left-hand side list represents a conjunction of
CIFF literals while the right-hand side list represents a
disjunction of CIFF atoms.
\end{itemize}

\noindent Equality/disequality atoms are defined via {\tt=,
$\backslash$==} and constraint atoms are defined via {\tt \#=,
\#$\backslash$=, \#<, \#=<, \#>, \#>=}\footnote{Note that, whenever possible,
disequalities in the system are managed through the operator
{$\backslash$==} rather than in the corresponding (and less efficient) implicative form.}. 
Finally, negative literals are of the form {\tt not(Atom)} where
{\tt Atom} is an ordinary atom.

\noindent All the clauses defining the same predicate (here
 a predicate is identified by its name plus its arity) are preprocessed by the
system in order to build the internal representation (an
iff-definition). Each iff-definition is asserted in the Prolog
global state in order to retrieve such information, when needed
during a CIFF derivation, in a simple and efficient way.

\noindent The CIFF proof rules are implemented in \CIFFQ{} as Prolog
clauses defining {\tt sat(+State, -Answer)}, where {\tt State}
represents the current selected CIFF node.

\noindent {\tt State} is initialized to the internal representation
of the {\tt Query} plus all the integrity constraints in (all files
in) the {\tt ALP} argument.

\noindent Throughout the computation {\tt State} is defined
as:\medskip

\begin{tt}
    state(Diseqs,CLPStore,Imps,Atoms,Abds,Disjs)\footnote{The representation of the current node,
    in the real code, needs some further elements dropped here for simplicity.}
\end{tt}\medskip

\noindent where the aggregation of the arguments represent a CIFF
 node. {\tt Diseqs} represents the set of CIFF disequalities, {\tt
CLPStore} represents the current finite domain constraint store,
{\tt Imps} the set implications, {\tt Atoms} the set of defined
atoms, {\tt Abds} the set of abduced atoms and finally {\tt Disjs}
is the set of disjunctive CIFF conjuncts in the node.

\medskip
\noindent The predicate {\tt sat} calls itself recursively until no
more rules can be applied to the current {\tt State}, thus
instantiating the {\tt Answer}.

\noindent Finally a note on the implemented CIFF selection function.
We use a classical Prolog-like selection function, i.e. we always
select the left-most CIFF node in a CIFF formula. It is not a
\emph{fair} selection function in the sense that it does not ensure
completeness (see Section \ref{sec:soundness} for further details),
but it has been found as the only possible practical choice in terms
of efficiency. Without entering in technical details, this is mostly
because, fixing the choice of the selected node in a CIFF formula as
the left-most CIFF node, we can directly take advantage of the
Prolog backtracking mechanism in order to switch to another CIFF
node in case of failure.

\noindent Concerning the order of selection of the proof rules in a
CIFF node, this is determined by the order of the {\tt sat} clauses.
If a {\tt sat} clause defining a CIFF proof rule, e.g.
\textbf{Unfolding atoms (R1)}, is placed before the {\tt sat} clause
defining e.g. \textbf{Propagation (R3)}, then the system tries first
to find a rule input for \textbf{R1} and, only if no such rule input
can be found, then the system tries \textbf{R3}.

\medskip
\noindent 
Below we sketch the most important  techniques used to make the
\CIFFSys{} an efficient abductive system. For further details on
these topics, please refer to \cite{CIFFThesis}.\medskip

\noindent \textbf{Managing variables and equalities.} Variables play
a fundamental role in nodes in CIFF: they can be either universally
quantified or existentially quantified. Universally quantified
variables can appear only in implications (which define their
scope). Existentially quantified variables can appear in any element
of the node, with scope the entire node. In the system the CIFF
variables are Prolog variables, but to distinguish at run-time
existentially quantified and universally quantified variables we use
the Prolog facility of attribute variables  \cite{attributedvar},
associating to each existentially quantified variable an {\tt
existential} attribute. Moreover, whenever possible, we use the
unification of Prolog for managing equality rewriting and
substitutions, but we also implemented the Martelli-Montanari
unification algorithm \cite{unification} for managing, in
particular, equality rewriting and substitutions involving
universally quantified variables.

\medskip
\noindent Many CIFF proof rules, for example, \textbf{Propagation
(R1)} and \textbf{Unfolding (R2, R3)} rules, typically need to be
followed by a set of \textbf{Equality rewriting (R8, R9)} and
\textbf{Substitution (R10, R11)} rules. In the \CIFFSys, these
\apici{equality} rules are not treated at the same level of the
other main proof rules, but rather they have been integrated within
them in order to improve efficiency. In particular rules
\textbf{R8}, \textbf{R9}, \textbf{R10}, \textbf{R11} are applied
transparently to the user (i.e. they are not defined as {\tt sat}
clauses) at the very end of the other proof rules, e.g. \textbf{R1},
\textbf{R2} and \textbf{R3}.

\medskip
\noindent \textbf{Loop management.} Recall that in the definition of
a CIFF derivation (Definition \ref{derivation}), we avoid repeated
applications of certain proof rules. In the \CIFFSys{} this
requirement is dealt with through a non-straightforward loop
management which is designed to avoid repetitive application of CIFF
proof rules, in particular \textbf{Propagation (R3)} and
\textbf{Factoring (R5)}, to the same rule input. Obviously, in order
to manage even small-medium size problems, loop management needs to
be efficient. We do not enter in details here, but just give a hint
of the technique. Loop management is done by enumerating univocally
each potential rule input component for \textbf{R3} and \textbf{R5}
(e.g. implications for \textbf{Propagation} and abducibles for
\textbf{Factoring}) in a CIFF node, maintaining them sequentially
ordered throughout the computation. Then, we can
(non-straightforwardly) avoid loops, applying proof rules
\textbf{R3} and \textbf{R5} to appropriate rule inputs, following
the order given by the enumeration.

\medskip
\noindent The loop management required in a CIFF derivation for
\textbf{Equality} and \textbf{Substitution} rules is, instead,
obtained (almost) for-free due to the integration of those proof
rules in the other main proof rules as discussed above.

\medskip
\noindent \textbf{Constraint solving.} Interfacing efficiently the
\CIFFSys{} with the underlying SICStus CLPFD solver is fundamental
for performance purposes. Despite a clear interface made available
by the Prolog platform, the main problem in the interaction with the
solver is that the solver binds variables to numbers when checking
the satisfiability of the current {\tt CLPstore} (i.e. when the
\textbf{Constraint Solving (R7)} rule is applied), while we want to
be able to return non-ground answers. The solution adopted in the
\CIFFSys{} tackles this problem through an algorithm which allows,
when needed, to check the satisfiability of the {\tt CLPstore} as
usual and then restores the non-ground values via a forced
backtracking.

\medskip
\noindent \textbf{Groundable integrity constraints}\label{sec:gics}
The main source of inefficiency in a CIFF computation is probably
represented by integrity constraints. The main problem is the
presence of universally quantified variables which potentially lead,
through the \textbf{Propagation} rule, to a new implication in a
CIFF node for each propagated variable instance. It is worth
noting that even in a small/medium size CIFF application, the
number of such implications resulting from integrity constraints
easily grows, thus representing the main computational bottleneck.

\noindent
To deal with this, we have incorporated within CIFF a specialized algorithm
that can be applied to a wide class of
integrity constraints, called \emph{groundable integrity
constraints}.
Intuitively, an integrity constraint $I$ is \emph{groundable} if the
set of implications obtained through the exhaustive application of
CIFF proof rules (in particular \textbf{Unfolding in implications}
and \textbf{Propagation}) on $I$ is \apici{expected to become
ground} at run-time. For example, consider an integrity constraint
of the form

$p(X),q(Y) \rightarrow r(X,Y)$

\noindent where $p$ and $q$ are both defined through a set of $N$
and $M$ ground facts respectively. Intuitively, the exhaustive
application of \textbf{Unfolding in implications} gives rise, at
run-time, to a set of $N*M$ implications which become ground after
the application of the substitutions on $X$ and $Y$. This type of
integrity constraint is included in the class of \emph{groundable
integrity constraints} which is formally defined in
\cite{CIFFThesis} together with the details of an algorithm for
managing it. This algorithm handles most of the operations on
groundable integrity constraints in the Prolog global state, via a
non-straightforward combination of assertions/retractions of the
(partial) instances of the groundable integrity constraints. The
system checks automatically, in the preprocessing phase, whether an
integrity constraint is a groundable integrity constraint and it
prepares all the needed data-structures. This feature significantly
boosts the performance of the system because firstly the operations
on implications performed in the Prolog global state are much faster
than the operations performed in a CIFF node in the usual way, and
secondly, the absence of a large set of implications in a node
boosts also the application of the proof rules to the other
elements. 

%

\begin{example}
\noindent The following is an example of groundable integrity
constraint:

\begin{verbatim}
  [q(R,C)] implies [p(R,C)].
\end{verbatim}

\noindent where {\tt q} is an abducible predicate. Indeed, for all
the concrete ground instances of $q$ which are abduced during a CIFF
derivation, the above integrity constraint gives rise to a set of
ground implication. Note that the class of groundable integrity
constraints includes integrity constraints containing abducibles in
their bodies because the algorithm also manages the cases in which
such abducibles are propagated to an abducible atom containing
existentially quantified variables.

\end{example}
\begin{example}
\noindent The following is an example of an integrity constraints
which is not groundable:

\begin{verbatim}
  [p(X)] implies [false].
\end{verbatim}

\noindent where a clause defining {\tt p(X)} is:

\begin{verbatim}
  p(Y).
\end{verbatim}

\noindent The problem in this case is given by the variable {\tt X}
in the body of the integrity constraint: unfolding {\tt p(X)} we
will obtain {\tt X = Y} and there is no way for {\tt Y} to be
grounded.
\end{example}

\def\Sciff{${\cal S}CIFF$}

\section{Related Work, Comparison and Experiments}
\label{cap4:secComparison}
\noindent
There is a huge literature on abductive logic programming
with and without constraints, see for example
\cite{abdsurvey93,abdsurvey98,abdsurvey01,km90,km91,ACLP-JLP00,pereira91,eshkow89,SLDNFA92,SLDNFA97,Asystemthesis,Asystem1,iff97,iffnaf,abdual1,linyou02,alias03,abdchr2,hyprolog}.
The closest systems to CIFF are the \Asys{} \cite{Asystemthesis} and \Sciff{}
\cite{sciff}.
The latter has also been developed as an extension of the IFF proof procedure to handle
numerical constraints as in CLP, but with focus on the specification and verification of
interactions in open agent societies. The main features of \Sciff{} are the support of dynamical happening of events
during computations, universally quantified variables in abducibles, the concept of fulfilment and violation of
expectations, given a set of events, and integrity constraints of a specialised
form which requires to include in their body at least one specific social construct
(an event or an expectation). Instead, CIFF is intended as a general purpose abductive proof procedure,
keeping the spirit of the original IFF proof procedure and conservatively adding numerical constraints.

\medskip
\noindent
The \Asys,
as remarked in \cite{Asystemthesis}, is a combination of three existing
abductive proof procedures, namely the IFF proof procedure
\cite{iff97}, the ACLP proof procedure \cite{ACLP-JLP00} and, most
importantly, the SLDNFA proof procedure \cite{SLDNFA97}, of which the
\Asys{} is a direct descendant.
The \Asys{} is the
state-of-the-art of abductive logic programming with constraints,
borrowing the most interesting features from the above cited proof
procedures.
In Section \ref{cap4:secCompASys} we give a detailed comparison between CIFF and
the \Asys.

\medskip
\noindent
Many approaches to abductive logic programming \cite{km90,km91,ACLP-JLP00,linyou02}
rely upon the stable models semantics
\cite{GelLif88} and its extensions.  Answer Set Programming (ASP) \cite{BarGel94}
is a logic programming based paradigm for computing stable models and answer set semantics.
The comparison of CIFF with
the two dominant answer set solvers,
DLV \cite{dlv1} and SMODELS \cite{smodels1}, is discussed
in Section \ref{cap4:secCompASP}.

\medskip
\noindent In Section \ref{cap4:secExperiments}, we present some
experimental results on concrete examples and in comparison with the
\Asys{} and the aforementioned answer set solvers. Note that
\cite{hyprolog} gives an extensive experimental comparison between
Hyprolog, another relevant system for abductive logic programming,
and CIFF, some ASP systems and the \Asys. Whereas CIFF is a
meta-interpreter, Hyprolog avoids meta-interpretation by directly
extending Prolog to incorporate abduction and constraint handling
\`a la CHR \cite{CHR}. However, Hyprolog
has restrictions on the use of negation, as mentioned in \cite{hyprolog}.

\medskip
\noindent Finally, in Section \ref{cap4:secTableaux} we give a
comparison with analytic tableaux-based methods.

\subsection{Comparison with \Asys}\label{cap4:secCompASys}
%
\noindent The \Asys{} and CIFF share many common points. They both
rely upon the three-valued completion semantics and their
computational schemas are both based on rewrite (proof) rules.
Moreover, both systems are implemented under SICStus Prolog and
 the syntax of the input programs is very similar. In both systems
much effort has been done, though adopting different solutions,
for obtaining considerable efficiency, by
exploiting the data structures and the services available in a
modern Prolog platform such as SICStus.
However there are also some important differences.

\medskip
\noindent \textbf{Treatement of Integrity Constraints} - The
\Asys{} framework requires that integrity constraints are in denial form.
Logically, implicative integrity constraints can be written in denial form, since
  \[ (B \rightarrow H) \equiv ((B \wedge \NAF H) \rightarrow \bot).\]
However, the operational treatement of the two representations of integrity constraints is
rather different in CIFF and in the \Asys. For example, given a CIFF integrity constraint
  \[ a \rightarrow b \]
(where $a$ and $b$ are abducibles) and an empty query, CIFF computes the empty set of abducibles,
whereas, given the equivalent denial
  \[ a \wedge \NAF b \rightarrow \bot \]
and the same query, the \Asys{} computes two alternative answers:
the empty set of abducibles and $\{b\}$. Indeed, assuming $b$ renders the original implication true.
However, in some applications this treatment leads to unintuitive behaviours.
For example, if $a$ is {\em alarm\_sounds} and $b$ is {\em evacuate}, then,
with the \Asys,
{\em evacuate} is a possible answer independently of whether
{\em alarm\_sounds} has been observed or not. This and other examples are discussed in
\cite{iffnaf}.

\medskip
\noindent \textbf{Negation in implications/denials} - The presence
of a negative literal ($\NAF A$) in the body of an implication is
handled by CIFF through a \textbf{Negation rewriting} rule
which moves $A$ to the head of
the implication. The \Asys{}, instead, manages such negations with a rule similar to a \textbf{Case Analysis} rule.
  That is, it creates a two-terms disjunction with a disjunct containing $A$ and the other disjunct containing ($\NAF A$) in conjunction with
  the rest of the original implication. This is exactly what CIFF does in the \textbf{Case
Analysis for equalities (R12)} and
  \textbf{Case Analysis for constraints (R6)} rules. However, as noted also in \cite{iffthesis}, applying a \textbf{Case Analysis} rule
  to a defined/abducible atom $A$, is not in the spirit of a {\em three-valued} semantics approach.
  This is the reason why in CIFF \textbf{Case Analysis}
  is used only for equalities and constraints, whose semantics
  is two-valued.

\subsection{Comparison with Answer Set
Programming}\label{cap4:secCompASP} \noindent Answer Set
Programming (ASP) (see, e.g. \cite{marek99,baralbook,BarGel94}) and
Abductive Logic Programming with
Constraints (ALPC) are strongly interconnected mechanisms for representing
knowledge and reasoning. This
interconnection arises at first glance, just noting that ASP is
based on the Answer Set Semantics \cite{AnswerSets}, an
\apici{evolution} of the stable models semantics \cite{GelLif88}
(which in turn is used as the core semantics for many abductive
proof procedures, e.g. \cite{ACLP-JLP00,km91,linyou02}) and that
abduction can be modeled in ASP, as shown e.g. in \cite{bonatti02}.

\medskip
\noindent Nevertheless, ASP and ALPC show important differences
which we briefly discuss here, assuming the reader has
some familiarity with ASP.

\medskip
\noindent The ASP framework is based upon some concrete assumptions.
In particular ASP relies upon
programs with finite Herbrand Universe
This assumption
has a high impact on the computational model and, hence, on the implemented answer set solvers.

\noindent The computational model of ASP,
relying upon programs with a finite Herbrand Universe, shares many common
points with typical constraint solving algorithms and it is very
distinct from the classic computational model of logic programming
(mostly used in ALPC and also in CIFF). For an excellent comparison
of the two computational models, see \cite{marek99}.

\noindent Directly from the above observations, the implemented
answer set solvers benefit from a number of features which have
made them popular tools for knowledge representation and reasoning:
\emph{completeness}, \emph{termination} and \emph{efficiency}.

\noindent Completeness and termination follows directly from the
assumption that the Herbrand universe of a program is finite. 

\noindent The idea of applying constraint solving techniques in the
computational model, together with hardware improvements, makes it
possible to have also efficient answer set solvers, and, indeed,
state-of-the-art solvers are able to handle hundreds of thousands of
ground Herbrand terms in acceptable times. This is sufficient for many
medium to large size applications.

\medskip
\noindent However, the ASP assumptions also introduce some important
limitations on the expressiveness of the framework. Even if many
application domains can be modeled through ASP, there are some
applications which need the possibility of introducing non-ground
terms. The web sites repairing example described in Section
\ref{ex:repair} below is one such (simple) application which is
being further investigated \cite{WWWCIF,WWV08CIF}. Moreover, there are
applications which can be effectively modeled in ASP, but for which
non-ground answers could be more suitable. Consider, for example, a
planning application where we search for a plan to solve a goal $G$
by time $T = 5$. Assume that a certain action $A$ solves the goal.
In a plan obtained from an answer set solver the action $A$ will be
bound to a ground time, for example $4$ or $3$. However, it might be
preferable to have a more general plan with $A$ associated with a
non-ground time $TA$ together with the constraint $TA \leq 5$.
Obviously, this is just a hint of a planning framework which is
outside the scope of this paper. Work focused on these topics
include, for example, \cite{planpart}, and part of the SOCS European
Project \cite{SOCS}.

\medskip
\noindent
To illustrate the main conceptual differences when programming applications
in ASP and CIFF, let us consider the well-known N-queens domain, where N
queens have to be placed on an N*N board in such a way that for no
pair of queens $Q_i$ and $Q_j$, $Q_i$ and $Q_j$ are in the same row
or in the same column or in the same diagonal.

\noindent We represent the problem in CIFF as follows ($N$ is a
placeholder for a natural number).

\[\begin{array}{l@{\quad}l}
\Prog: & exists\_q(R) \leftarrow q\_domain(R) \wedge q\_domain(C) \wedge q\_pos(R,C)\\
& q\_domain(R) \leftarrow R \geq 1 \wedge  R \leq N\\
& safe(R1,C1,R2,C2) \leftarrow C1 \neq C2 \wedge  \: (R1 + C1 \neq R2 + C2) \wedge  \\
& \qquad \qquad \qquad \qquad \qquad \: \: \: (C1 - R1 \neq C2 - R2)\\[1pt]

\Abd: & \{ q\_pos \}\\[1pt]
\IC: & q\_pos(R1,C1) \wedge  q\_pos(R2,C2) \wedge  R1 \neq R2 \rightarrow safe(R1,C1,R2,C2)\\[2pt]
Q: & exists\_q(1) \wedge  \ldots \wedge  exists\_q(N)\\[2pt]
\end{array}\]

\noindent The CIFF specification of the problem is very compact.  A
CIFF computation for the query $Q$ proceeds as follows (we abstract
away from the concrete CIFF selection function). Each $exists\_q(R)$
atom in the query (where $R$ is one of the $N$ integer values
between $1$ and $N$) is unfolded giving rise to three atoms:
$q\_domain(R)$, $q\_domain(C)$ and the abducible $q\_pos(R,C)$. The
first two atoms are in turn unfolded populating the CIFF node with
the finite-domain constraints:

\[ R \geq 1, R \leq N, \qquad \qquad C \geq 1, C \leq N\]

\noindent which will be evaluated by the constraint solver. Note
that the constraints concerning $R$ are obviously \emph{ground},
while the constraints concerning $C$ are not \emph{ground} due to
the presence of $C$.

\noindent The third atom $q\_pos(R,C)$ is instead an abducible
non-ground atom (due to the presence of the constraint variable
$C$).

\noindent Assuming that all the unfolding, the equality rewriting
and the substitutions have been done, we will obtain a node with the
following abducible atoms:

\[ q\_pos(1,C_1), \ldots, q\_pos(N,C_N) \]

\noindent Each pair of these has to be propagated to the integrity
constraint firing $N^2$ non-ground instances of the $safe$ atom. The
condition $R1 \neq R2$ in the body of the integrity constraint in
$IC$ avoids to propagate twice the same abducible, i.e. it avoids to
have an instance like $safe(R_1,C_1,R_1,C_1)$.

\noindent At this point the $safe$ atoms are unfolded, resulting in
the whole set of non-ground finite-domain constraints needed to
ensure correct positioning of the queens. Finally, this set, once
the solver checks its satisfiability, is returned as part of the
extracted answer. The extracted answer \apici{contains} all the
possible solutions: the corresponding ground answers identifying the
concrete positions of the queens can be obtained performing a
\emph{labeling} on the constraint variables (the \CIFFSys{}
automatically performs the final labeling if the user wishes it).

\medskip
\noindent Consider now the following ASP representation
\footnote{We
choose the DLV representation, borrowed from

 \qquad \qquad \url{http://www.dbai.tuwien.ac.at/proj/dlv/tutorial/},

because it is the closest representation to ours and
we can easily highlight the differences. For the same reason we
present the DLV specification as a set of ALPC integrity
constraints: DLV syntax is slightly different.}:

\[\begin{array}{l@{\quad}l}
& row(1)\\
& \ldots\\
& row(N)\\[3pt]
& row(R) \rightarrow  q\_pos(R,1) \vee \ldots \vee q\_pos(R,N)\\[2pt]
& q\_pos(R1,C) \wedge  q\_pos(R2,C) \wedge  R1 \neq R2 \rightarrow \bot\\[2pt]
& q\_pos(R1,C1) \wedge  q\_pos(R2,C2) \wedge row(R) \wedge R2=R1+R \wedge  C1=C2+R \rightarrow \bot\\[2pt]
& q\_pos(R1,C1) \wedge  q\_pos(R2,C2) \wedge  row(R) \wedge  R2=R1+R \wedge  C2=C1+R \rightarrow \bot\\[2pt]
\end{array}\]

\noindent Also in this case all the possible solutions are returned
by the answer set solvers, even if enumerating them in a ground
form.

\noindent Abstracting away from syntactical differences, there
is an important difference between the two specifications. The CIFF
specification takes advantage of the constraint solver because it
delegates the constraints on the variables inside the clause
concerning the $safe$ predicate as informally described above.
Conversely, in an ASP computation, the conditions on the queen
positions are checked locally, resulting in a huge set of
\emph{groundable} integrity constraints, each one containing a
ground pair of queen positions.

\noindent As expected (and as shown in Section \ref{cap4:secNqueens}
below), delegating the checks to a finite-domain constraint solver
results in performances an order of magnitude faster than any answer set solver. Note that the ASP community is aware of this problem
and recently some work has been initiated on integrating ASP with constraint solvers,
in an effort to reduce the grounding size and speed computation
(e.g., \cite{BBG05,MG08}), but for limited forms of constraints and
restricted combinations
of logic programs and constraints.

\subsection{Experimental Results}\label{cap4:secExperiments}
\noindent In this section, we show some experimental results
obtained running two of the most typical benchmark examples,
namely the \emph{N-Queens} problem and the \emph{graph coloring}
problem. We also present a simple instance of a web sites repairing
framework which could be used with CIFF. Note that we focus our
experimental evaluation on examples where abduction benefits
from constraint solving, in
order to illustrate the main innovative feature of CIFF with respect to its
predecessor IFF, as well as related systems (ALP solvers and \Asys).

\noindent In this performance comparison we restricted our attention
to three systems: the \Asys{} \cite{Asystemthesis} and two
state-of-the-art answer set solvers, namely the DLV system
\cite{dlv1} and SMODELS \cite{smodels1}.

\noindent All the tests have been run on a Fedora Core 5 Linux
machine equipped with a 2.4 Ghz PENTIUM 4 - 1Gb DDR Ram. The SICStus
Prolog version used throughout the tests is the 3.12.2 version. All
execution times are expressed in seconds (``---'' means that the
system was still running after 10 minutes). In all examples, unless
otherwise specified, the \CIFFSys{} query is the empty list {\tt[]}
representing $true$ and the algorithm groundable integrity
constraint is activated.
In each experiment, the formalisation of the problems
are taken
from \url{http://www.dbai.tuwien.ac.at/proj/dlv/tutorial/} for DLV,
from \url{http://www.baral.us/code/smodels/} for SMODELS, and from
\cite{Asystemthesis} for \Asys.

\subsubsection{The N-Queens problem}\label{cap4:secNqueens}
We recall the N-Queens, already seen in Section
\ref{cap4:secCompASP}: N queens have to be placed on an N*N board in
such a way that for no pair of queens $Q_i$ and $Q_j$, $Q_i$ and
$Q_j$ are in the same row or in the same column or in the same
diagonal.

\medskip
\noindent The \CIFFSys{} formalization (\textbf{CIFF (1)}) of this
problem is very simple (the query is a conjunction of $N$ {\tt
exists\_q(R)} where each $R$ is a natural number, distinct from each
other, in $[1,N]$):

\begin{verbatim}
  %%% CIFF (1)
  %%% ABDUCIBLES
  abducible(q_pos(_,_)).

  %%% CLAUSES
  q_domain(R)  :- R #>= 1, R #=< N.
    %%% N must be an integer in real code!

  exists_q(R) :- q_domain(R),q_pos(R,C),q_domain(C).

  safe(R1,C1,R2,C2) :- C1#\=C2, R1+C1#\=R2+C2, C1-R1#\=C2-R2.

  %%% INTEGRITY CONSTRAINTS
  [q_pos(R1,C1),q_pos(R2,C2),R1#\=R2] implies [safe(R1,C1,R2,C2)].
\end{verbatim}

\noindent We also show another CIFF formalization which is a direct
translation of the DLV formalization. Here, the checks on the queen
position conditions, are made locally in each groundable integrity
constraint instance and they are not delegated to the constraint
solver. In these programs, {\tt abs} is the absolute value function.

\medskip
\noindent The DLV translation (\textbf{CIFF (2)}) is very similar to
the (\textbf{CIFF (1)}) formalization and the query is the same. But
in this case the conditions on the queen positions is done locally
in the body of the integrity constraints\footnote{The concrete CIFF
syntax differs a bit from that of the program shown in Section
\ref{cap4:secCompASP}. The conditions which avoid to place two
queens in the same diagonal are integrated in a single integrity
constraint, taking advantage of the {\tt -} and {\tt abs} functions
of the constraint solver: the DLV system does not allow to express
such functions. The straight DLV translation with two integrity
constraints runs a bit slower in CIFF, as expected.}.

\begin{verbatim}
  %%% CIFF (2)
  %%% DLV translation
  %%% ABDUCIBLES
  abducible(q_pos(_,_)).

  %%% CLAUSES
  row(1).
  ...
  row(N).

  %%% INTEGRITY CONSTRAINTS
  [row(R)] implies [q_pos(R,1), ..., q_pos(R,N)].
    %%% N must be an integer in real code!

  [q_pos(R1,C),q_pos(R2,C),R1\==R2] implies [false].

  [q_pos(R1,C1),q_pos(R2,C2),R1\==R2,(abs(R1-R2)#=abs(C1-C2))]
    implies [false].
\end{verbatim}

\medskip
\noindent In Table \ref{nqueensresults}, we show the results for the first solution found.
In the tables, we denote the \Asys{} as \textbf{ASYS} and the
SMODELS as \textbf{SM}.

\begin{table}[ht]
\caption{{\tt N-Queens} results (first
solution)}\label{nqueensresults}
\begin{tabular}{lrrrrr}
\hline\hline

    \textbf{Queens}  & \textbf{CIFF (1)} & \textbf{CIFF (2)}  & \textbf{ASYS}  & \textbf{SM}   & \textbf{DLV}\\
    \hline
    {\tt n} = 4      & 0.01                  & 0.02           &  0.01              & 0.01               & 0.01\\
    {\tt n} = 6      & 0.01                  & 0.21           &  0.01              & 0.01               & 0.01\\
    {\tt n} = 8      & 0.03                  & 1.29           &  0.03              & 0.01               & 0.01\\
    {\tt n} = 12     & 0.05                  & 5.98           &  0.05              & 0.01               & 0.01\\
    {\tt n} = 16     & 0.09                  & 410.33         &  0.07              & 0.36               & 0.61\\
    {\tt n} = 24     & 0.20                  & ---            &  0.17              & 4.88                & 5.44\\
    {\tt n} = 28     & 0.29                  & ---            &  0.27              & 55.32                & 35.17\\
    {\tt n} = 32     & 0.37                  & ---            &  0.32              & ---             & ---\\
    {\tt n} = 64     & 1.62                  & ---            &  1.52              & ---              & ---\\
    {\tt n} = 100    & 4.55                  & ---            &  4.24              & ---              & ---\\

\hline\hline
\end{tabular}
\end{table}

\noindent All systems return all the correct solutions, but we do
not show the times for all solutions because the number of possible
solutions is huge when $N$ grows.

\medskip
\noindent Only the \CIFFSys{} and the \Asys{}, through the use of
the finite domain constraint solver, can solve the problem, in a
reasonable time, for a high number of queens. Note also that the
\CIFFSys{} performances in the other \apici{answer set} variants of
the specification, i.e. \textbf{CIFF (2)}, is, as expected, worse in
comparison with the first one, i.e. \textbf{CIFF (1)}. 
However, we argue that, on the whole, the results show that the
system is able to handle a reasonable number of ground instances.

\subsubsection{The Graph Coloring problem}\label{cap4:Coloring}

\noindent The graph coloring problem can be defined as follows:
given a connected graph we want to color its nodes in a way that
each node does not have the color of any of its neighbors.

\noindent The \CIFFSys{} formalization is as follows (again, we omit
the domain-dependent definitions of any specific graph):

\begin{verbatim}
   %%% ABDUCIBLES
   abducible(abd_color(_,_)).

   %%% CLAUSES
   coloring(X) :- color(C),abd_color(X,C).

   %%% INTEGRITY CONSTRAINTS
   [vertex(X)] implies [coloring(X)].
   [edge(X,Y),abd_color(X,C),abd_color(Y,C)] implies [false].
\end{verbatim}

\noindent The results are the following, where {\tt Jean} and {\tt
Games} are two graph instances
  (up to a 120-nodes graph)\footnote{They are borrowed from
  {\tt http://mat.gsia.cmu.edu/COLOR/instances.html}.}:

\begin{table}[ht]
\caption{{\tt Graph coloring} results (first
solution).}\label{colorresults}
\begin{tabular}{lrrrrr}
\hline\hline

    \textbf{Nodes}  & \textbf{CIFF} & \textbf{CIFF (G)}  & \textbf{ASYS} & \textbf{SM}   & \textbf{DLV}\\
    \hline
    4                     &  0.09   &   0.01  &   0.01      & 0.01    & 0.01\\
    Jean                   &  ---   &   0.68  &   0.60      & 0.19    & 0.48\\
    Games                 &  ---   &   2.39   &   3.61     & 0.28    & 1.14\\

\hline\hline
\end{tabular}

\end{table}

\noindent As for the N-Queens problem all the systems return all the
solutions. Here answer set solvers have the best performances as
the constraint solver is not involved in the computation. However,
it is worth noting that performances of both the \Asys{} and the
\CIFFSys{}, when the algorithm for \emph{groundable} integrity
constraints is activated (second column), are encouraging, even if
the domain is a typical ASP application.

\subsubsection{Web Sites Repairing}\label{ex:repair}
\noindent The last example is a practical problem in which abduction
can be used effectively: checking and repairing links in a web site,
given the specification of the site via an abductive logic program
with constraints. This example, which follows the approach in
\cite{toniinfo}, is currently being formalized, expanded and
investigated \cite{WWWCIF,WWV08CIF,CIFFThesis}.

\noindent Consider a web site where a \emph{node} (representing a
web page) can be a \emph{book}, a \emph{review} or a \emph{library}.
A \emph{link} is a relation between two nodes. Nodes and links may
need to be added to guarantee some properties.

\begin{itemize}
  \item each node must not belong to more than one type, and
  \item each book must have at least a link to both a review
and a library.
\end{itemize}

\noindent We represent the addition of links and nodes as abducibles
and we impose that:

\begin{itemize}
  \item each abduced node must be distinct from each other node (either abduced or
  not),
  \item each abduced link must be distinct from each other link (either abduced or
  not),
\end{itemize}

\noindent The \CIFF{} formalization of this problem (together with a
simple web site instance) is the following:

\begin{verbatim}
 %%% ABDUCIBLES
 abducible(add_node(_,_)).
 abducible(add_link(_,_)).

 %%%CLAUSES
 is_node(N,T) :- node(N,T), node_type(T).
 is_node(N,T) :- add_node(N,T), node_type(T).
 node_type(lib).
 node_type(book).
 node_type(review).

 is_link(N1,N2) :- link(N1,N2), link_check(N1,N2).
 is_link(N1,N2) :- add_link(N1,N2), link_check(N1,N2).
 link_check(N1,N2) :- is_node(N1,_), is_node(N2,_), N1 \== N2.
 book_links(B) :- is_node(B,book), is_node(R,review), is_link(B,R),
                  is_node(L,lib), is_link(B,L).

 %%% INTEGRITY CONSTRAINTS
 [add_node(N,T1), node(N,T2)] implies [false].
 [add_link(N1,N2), link(N1,N2)] implies [false].
 [is_node(N,T1), is_node(N,T2), T1 \== T2] implies [false].
 [is_node(B,book)] implies [book_links(B)].

 %%%WEB SITE INSTANCE
 node(n1,book).
 node(n3,review).
 link(n1,n3).
\end{verbatim}

\noindent The \CIFFSys{} returns two answers representing correctly
the need of a new link between the \emph{book} {\tt n1} and a new
\emph{library node} {\tt L}. The first answer is:\vspace{-0.2cm}

\medskip
\begin{verbatim}
  [add_link(n1,L), add_node(L,lib)],   %%%ABDUCIBLES
  [L\==n3, L\==n1],                    %%%DISEQUALITIES
  []                                   %%%FD CONSTRAINTS
\end{verbatim}\vspace{-0.2cm}

\medskip
\noindent Note that in the answer it is included the fact that {\tt
L} must be a \emph{new} node, i.e. a node distinct from both {\tt
n1} and {\tt n3}.

\noindent The second answer is more complex:

\medskip
\begin{verbatim}
  [add_link(n1,L), add_node(L,lib),
   add_link(n1,R), add_node(R,review)],      %%%ABDUCIBLES
  [L\==n3, L\==n1, R\==n3, R\==n1, R\==L],   %%%DISEQUALITIES
  []                                         %%%FD CONSTRAINTS
\end{verbatim}\vspace{-0.2cm}

\medskip
\noindent In this case, the system also adds a \emph{new review}
node {\tt R} and provides the right links among the new nodes. Note
that, again, each node must be distinct from each other: this is
expressed through CIFF disequalities.

\noindent Correctly, no further answers are found and the system
terminates accordingly.

\noindent For this example we do not make a performance comparison
with other systems as both answer set solvers  and the \Asys{} seem
 unable to provide correct answers due to the presence of unbound
variables.

\subsection{Comparison with Analytic Tableaux}\label{cap4:secTableaux}


The overall framework of the CIFF procedure resembles the method of
analytic tableaux, which has been used mostly for deductive
inference in a range of different logics~\cite{tableau-handbook}. A
tableau proof proceeds by initializing a proof tree with a set of
formulas to which we then apply expansion rules, similar to those of
CIFF, until we reach an explicit contradiction on every branch. This
can be used to prove that a set of formulas $T$ is unsatisfiable or
that a formula $\varphi$ follows from a set $T$ (by adding the
complement of $\varphi$ to $T$ before expansion). There has been a
(very limited) amount of work on applying the tableau method to the
problem of abductive
inference~\cite{MayerPirriIGPL1993,AlisedaLlera1997,Klarman2008}.
The basic idea is that if an attempted proof of $T\models\varphi$
fails, then those branches that could not be closed can provide
hints as to what additional formulas would allow us to close all
branches. That is, we can compute an abductive answer for the query
$\varphi$ given the theory $T$ in this manner. While, in principle
it is possible to use such an approach, the search space would be
enormous. The rules of CIFF (which are more complicated and tailored
to specific cases than the rules of most tableau-based procedures)
have been specifically designed so as to avoid at least some of this
complexity and search for to abductive answers more directly. Most
work on tableau-based abduction has concentrated on (classical and
non-classical) propositional
logics~\cite{AlisedaLlera1997,Klarman2008}. The only work on
tableau-based abduction for first-order logic that we are aware of
does not focus on algorithmic issues~\cite{MayerPirriIGPL1993}. We
are also not aware of any major implementations of any of the
tableau-based procedures for abduction proposed in the literature.

\section{Conclusions}\label{cap4:secConclusions}
\noindent We have presented the CIFF proof procedure, a
step forward at both theoretical and implementative levels in the
field of abductive logic programming (with constraints).
CIFF is able to handle
variables in a non-straightforward way, and it is equipped with a
useful interface to a constraint solver.
We have proved that CIFF is sound with respect to the
three-valued completion semantics, and it enjoys some
completeness properties with respect to the same semantics.

\medskip
\noindent
In addition, we have described the \CIFFSys, a Prolog implementation
of the CIFF proof procedure. The \CIFFSys{} reaches good levels of
efficiency and flexibility and is comparable to other
state-of-the-art tools for knowledge representation and
reasoning. The system has been developed in SICStus Prolog, but recently ported
to SWI-Prolog \cite{SWIProlog}, the state-of-the-art open-source Prolog platform,
whose constraint solver is however less efficient than the one in SICStus.


\medskip
\noindent
We have developed an extension of CIFF incorporating a more sophisticated
form of integrity constraints, with negation as failure in their bodies. This extension
is inspired by \cite{iffnaf} and is described in \cite{CIFFThesis}.
Even though the current implementation supports this extended treatment of negation,
further work is needed to give it a formal foundation.

%
%
%

\medskip
\noindent At the implementative level, a main issue in CIFF is the
lack of a Graphical User Interface (GUI) which would improve its
usability: we hope to add it in the \CIFFSys{} 5 release.

\medskip
\noindent Other interesting features which are planned to be added
to the \CIFFSys{} 5 release, are the following.

\begin{itemize}
    \item Compatibility to the SICStus Prolog 4 release (which is
claimed to be much faster: a porting of the system will benefit at
once from this boost in performances).
    \item The possibility of invoking Prolog platform functions
    directly. We think that this would enhance performances and
    ease-of-programming in CIFF. However, some work has to be done
    in order to understand how to integrate them safely.
    \item Further improvements in the management of {\em groundable integrity
    constraints}.
    \item Further experimentations with other applications, for example planning.
\end{itemize}

Finally, we also plan to compare the CIFF system with tools in Potassco (the Potsdam Answer Set Solving Collection)~\footnote{{\tt http://potassco.sourceforge.net/}}, that incorporate efficient implementations of constraint solving within answer set programming.

\bigskip
\noindent
{\bf Acknowledgements}: We would like to thank Michael Gelfond and the anonymous reviewers for their
comments and suggestions. The work described in this paper has been
partially supported by European Commission FET Global Computing Initiative,
within the SOCS project (IST-2001-32530).

\bibliographystyle{acmtrans}
\bibliography{CIFF}

\vfill

\pagebreak

\appendix

\section{Proofs of CIFF results}\label{AppendixA}

\begin{proof*}[Proof of Proposition \ref{lemmasub1}]
\noindent To prove the first part of the proposition, we need the
semantics of the constraint solver while to prove the second part we
need the \emph{Clark Equality Theory (CET)}. Both are embedded in
our semantics ($\models_{3(\Re)}$) and we will write explicitly
$\models_{\Re}$ and $\models_{CET}$, respectively, instead of
$\models_{3(\Re)}$ where appropriate.

\medskip
%

\begin{enumerate}
  \item $\Gamma$ is the set of c-conjuncts in $N$, and this is a
successful node. Then the \textbf{Constraint solving} rule
\textbf{R7} cannot be applied to $N$. Thus, by the assumption of
having a sound and complete constraint solver, we have that $\Gamma$
is not an unsatisfiable set of constraints, i.e. we can always
obtain a ground substitution $\theta$ such that:

\[ \models_\Re \Gamma\theta \]

\noindent and so

\[\theta \models_{\Re} \Gamma.\]

  \item Let us consider $F = (E \cup DE)\theta$. Equalities in $E$ are of the form

\[ X_i = t_i  \qquad  (1 \leq i \leq n, n \geq 0) \]

\noindent where each $X_i$ is an existentially quantified variable
and $t_i$ is a term (containing neither universally quantified
variables nor $X_i$ itself). The scope of each variable in $E$ is
the whole CIFF node $N$ and each $X_i$ does not appear elsewhere in
the node due to the exhaustive application of the \textbf{Equality
rewriting in atoms} rule \textbf{R8}.

\medskip
\noindent The disequalities in $DE$ are of the form

\[ X_j = t_j \rightarrow \bot  \qquad  (n < j \leq m, m \geq 0) \]

\noindent where each $X_j$ is an existentially quantified variable
appearing also in $E$ (due to the \textbf{Substitution in atoms}
rule \textbf{R10}) and $t_j$ is a term not in the form of a
universally quantified variable.

\medskip
\noindent The ground substitution $\theta$ contains an assignment to
all the constraint variables occurring in $(E \cup DE)$. This is
because (i) all the equalities in $E$ are equalities over Herbrand
terms by definition and (ii) there is no CIFF disequality in $DE$ of
the form $X_i = t_i \rightarrow \bot$ where $X_i = t_i$ is a c-atom
because the \textbf{Case analysis for constraints} rule \textbf{R6}
replaced any such CIFF disequality with a c-conjunct of the form
$X_i \neq t_i$.

\noindent Note that also CIFF disequalities of the form $X = Y
\rightarrow \bot$ such that $X$ is a constraint variable and $Y$ is
not (or viceversa) are not a problem. This is because $X$ has been
substituted by a ground term $c$ by $\theta$ and there is no
equality of the form $Y = c$ in $E\theta$ because in that case also
$Y$ would be a constraint variable and that equality would belong to
$\Gamma$.

\medskip
\noindent Finally, the proposition is proven by finding a ground
substitution $\sigma$ such that $\models_{CET}  F\sigma$ and this
can be done following the proof in \cite{iffthesis}, as follows.

\medskip
\noindent First we assign a value to each existentially quantified
variable $X_j$ in $DE\theta$. We do this by using a fresh function
symbol $g_j$, i.e. the function symbol $g_j$ does not appear in the
CIFF branch whose leaf is $N$ (we assume here that we have an
infinite number of distinct function symbols in our language). Then
we choose a constant $c$ and we assign $g_j(c)$ to $X_j$. 
We define $G = F\sigma_{I}$ where $\sigma_{I}$ is the ground
substitution composed of the above assignments.

\medskip
\noindent The second step is to assign to each variable $X_i$ in
$(E\theta)\sigma_{I}$ its corresponding term $s_i = t_i\sigma_{I}$.

\medskip
\noindent Finally, for each remaining existentially quantified
variable, we use another fresh function and a constant $c$ to make
assignments as for what done for CIFF disequalities.

\medskip
\noindent The whole set of assignments so far obtained is the ground
substitution $\sigma$ which proves the proposition. This is because,
after $\theta\sigma$ has been applied, each equality originally in
$E$ is of the form $t = t$ and each CIFF disequality originally in
$DE$ is of the form $f(t) = g(t) \rightarrow \bot$ which are
obviously entailed by CET.

\noindent We have:

\[ \sigma \models_{3(\Re)} (E \cup DE)\theta\]

\noindent and thus, being $\theta \models_{\Re} \Gamma$, we have:

\[ \theta\sigma \models_{3(\Re)} \Gamma \cup E \cup DE \mathproofbox\]
\end{enumerate}

\end{proof*}

\begin{proof}[Proof of Proposition \ref{lemmasub2}]
Let us consider the set $\Delta\sigma'$. There can be existentially
quantified variables in $\Delta$ not assigned by $\sigma'$ because
they do not appear in $C$. Then it is enough to choose arbitrary
ground terms to assign to those variables to obtain a substitution
$\sigma$ such that $\sigma' \subseteq \sigma$, which proves the
proposition.
\end{proof}

\begin{proof}[Proof of Proposition \ref{lemmaeq}]
\noindent We prove the proposition considering each of the CIFF
proof rules in turn. Recall that, apart from the \textbf{Splitting}
rule, for each proof rule the set \Nodes{} of successor nodes of $N$
is a singleton, i.e. \Nodes{} $= \{ N' \}$.

\medskip
\noindent \textbf{R1 - Unfolding atoms}. This rule applies a
resolution step on a defined atom $p(\vec{t})$ in $N$ and its
iff-definition in $Th$:

\[ p(\vec{X}) \leftrightarrow (D_1\vee\cdots\vee D_n) \]

\noindent Hence, the atom $p(\vec{t})$ is replaced in $N'$ by

\[ (D_1\vee\cdots\vee D_n)[\vec{X}\!/\vec{t}] \]

\noindent The replacement is obviously equivalence preserving with
respect to $P$ and $\models_{3(\Re)}$.

\medskip
\noindent \textbf{R2 - Unfolding within implication}. This rule
resolves a defined atom $p(\vec{t})$ with its iff-definition in
$Th$:

\[ p(\vec{X}) \leftrightarrow (D_1\vee\cdots\vee D_n) \]

\noindent as in the previous rule. The result is a set of
implications in $N'$ replacing the original implication, each one
containing one of the disjuncts $D_i\theta$, with $1 \leq i \leq n$
where $\theta = [\vec{X}\!/\vec{t}]$. Without loss of generality,
suppose that the original implication is of the form

\[ ( p(\vec{t}[\vec{W},\vec{Y}]) \wedge R[\vec{W},\vec{Y}]) \rightarrow H[\vec{W},\vec{Y}]\]

\noindent where $R$ is a conjunction of literals and $H$ is a
disjunction of atoms. We use the notation $E[\vec{Y}]$ to say that
$\vec{Y}$ may occur in $E$ for a generic $E$. Suppose that all and
only the variables in $\vec{W}$ occur also in another
non-implicative CIFF conjunct (recall that in a CIFF node variables
appearing only within an implication are implicitly universally
quantified with scope the implication itself and variables appearing
outside an implication are existentially quantified with scope the
whole node). Making the quantification explicit, the implication
becomes:

\[ \exists \vec{W} \forall \vec{Y} ( p(\vec{t}[\vec{W},\vec{Y}]) \wedge R[\vec{W},\vec{Y}] \rightarrow H[\vec{W},\vec{Y}] )\]

\noindent  To simplify the presentation, in the following we assume
that $\vec{W}$ and $\vec{Y}$ may occur everywhere in the implication
without denoting it explicitly. Applying resolution we obtain:

\[ \exists \vec{W} \forall \vec{Y} ( (\exists \vec{Z}_1 D'_1\theta \vee\cdots\vee \exists \vec{Z}_n D'_n\theta) \wedge R \rightarrow H )\]

\noindent where each $D_i$ is of the form $\exists \vec{Z}_i D'_i$
and the vectors $\vec{Z}_i$ of existentially quantified variables
arise from the iff-definition. Thus we have:

\[\begin{array}{l@{\quad}l}

 \exists \vec{W}\forall \vec{Y} ( (\exists \vec{Z}_1 (D'_1\theta)
\vee\cdots\vee \exists \vec{Z}_n (D'_n\theta)) \wedge R \rightarrow
H
) & \equiv\\[2pt]

 \exists \vec{W} \forall \vec{Y} ( \neg (\exists \vec{Z}_1
 (D'_1\theta)
\vee\cdots\vee \exists \vec{Z}_n (D'_n\theta)) \vee \neg R \vee H
) & \equiv\\[2pt]

 \exists \vec{W} \forall \vec{Y} ( (\neg (\exists \vec{Z}_1
 (D'_1\theta)) \wedge
\cdots\wedge \neg (\exists \vec{Z}_n (D'_n\theta))) \vee \neg R \vee
H
) & \equiv\\[2pt]

 \exists \vec{W} \forall \vec{Y} ( (\neg (\exists \vec{Z}_1
 (D'_1\theta)) \vee \neg R \vee H) \wedge
\cdots\wedge (\neg (\exists \vec{Z}_n (D'_n\theta)) \vee \neg R \vee
H)
) & \equiv\\[2pt]

 \exists \vec{W} (\forall \vec{Y} (\neg (\exists \vec{Z}_1
 (D'_1\theta)) \vee \neg R \vee H) \wedge
\cdots\wedge \forall \vec{Y} (\neg (\exists \vec{Z}_n (D'_n\theta))
\vee \neg R \vee H))
 & \equiv\\[2pt]

 \exists \vec{W} (\forall \vec{Y},\vec{Z}_1 (\neg D'_1\theta \vee \neg R \vee H) \wedge
\cdots\wedge \forall \vec{Y},\vec{Z}_n (\neg D'_n\theta \vee \neg R
\vee H))
 & \equiv\\[2pt]

 \exists \vec{W} (\forall \vec{Y},\vec{Z}_1 (D'_1\theta \wedge R \rightarrow H) \wedge
\cdots\wedge \forall \vec{Y},\vec{Z}_n (D'_n\theta \wedge R
\rightarrow H))
 & \\
\end{array}\]

\noindent Note that the variables $\vec{Z}_i$ in the new
implications are universally quantified with scope the implication
in which they occur. So with our convention for implicit
quantification, the last sentence is:

\[\begin{array}{l@{\quad}l}
 (D_1\theta \wedge R \rightarrow H) \wedge
\cdots\wedge (D_n\theta \wedge R \rightarrow H).
 & \\
\end{array}\]

\medskip
\noindent \textbf{R3 - Propagation}. This rule uses an atomic CIFF
conjunct $p(\vec{s})$ and an atom $p(\vec{t})$ within an implication
of the form $(p(\vec{t})\wedge B) \rightarrow H$ and it adds in $N'$
an implication of the form:

\[ \vec{t}=\vec{s} \wedge B \rightarrow H \]

\noindent It is obvious that, due to the fact that the second
implication is a consequence of the CIFF conjunct and the
implication and both remain in $N'$, the \textbf{Propagation} rule
is equivalence preserving.

\medskip
\noindent \textbf{R4 - Splitting}. This rule uses a disjunctive CIFF
conjunct of the form $D = D_1 \vee \ldots \vee D_k$ and builds a set
of CIFF successor nodes \Nodes $ = \{ \sequenza{N}{k} \}$ such that
in each $N_i$ the conjunct $D$ is replaced by $D_i$.

\noindent It is obvious that the \textbf{Splitting} rule is
equivalence preserving because it is an operation of disjunctive
distribution over a conjunction, i.e. is a case of the tautology:

\[ A \wedge (D_1 \vee \ldots \vee D_k) \equiv (A \vee D_1) \wedge \ldots \wedge (A \vee D_k) \]

\medskip
\noindent \textbf{R5 - Factoring}. This rule uses two atomic CIFF
conjuncts of the form $p(\vec{t})$ and $p(\vec{s})$ and it replaces
them in $N'$ by a disjunction of the form:

\[ (p(\vec{s}) \wedge p(\vec{t}) \wedge (\vec{t}=\vec{s} \rightarrow \bot)) \vee (p(\vec{t}) \wedge \vec{t}=\vec{s})  \]

\noindent To show that the rule is equivalence preserving, consider
the tautology

\[ (\vec{t}=\vec{s} \rightarrow \bot) \vee \vec{t}=\vec{s}  \]

\noindent We have that

\[\begin{array}{l@{\quad}l}

p(\vec{t}) \wedge p(\vec{s}) & \equiv\\
p(\vec{t}) \wedge p(\vec{s}) \wedge ((\vec{t}=\vec{s} \rightarrow \bot) \vee \vec{t}=\vec{s}) & \equiv\\
(p(\vec{t}) \wedge p(\vec{s}) \wedge (\vec{t}=\vec{s} \rightarrow \bot)) \vee (p(\vec{t}) \wedge p(\vec{s}) \wedge \vec{t}=\vec{s}) & \equiv\\
(p(\vec{t}) \wedge p(\vec{s}) \wedge (\vec{t}=\vec{s} \rightarrow \bot)) \vee (p(\vec{t}) \wedge p(\vec{t}) \wedge \vec{t}=\vec{s}) & \equiv\\
(p(\vec{t}) \wedge p(\vec{s}) \wedge (\vec{t}=\vec{s} \rightarrow \bot)) \vee (p(\vec{t}) \wedge \vec{t}=\vec{s}) & \\
\end{array}\]


\medskip
\noindent \textbf{R6 - Case Analysis for constraints}.

\noindent Recall that variables in $Con$ are all existentially
quantified and that the constraint domain is assumed to be closed
under complement, i.e. the complement $\overline{Con}$ of a
constraint atom $Con$ is a constraint atom.

\[\begin{array}{l@{\quad}l}

(Con \wedge A) \rightarrow B& \equiv\\
Con \rightarrow (A \rightarrow B)& \equiv\\
(Con \rightarrow Con) \wedge (Con \rightarrow (A \rightarrow B))& \equiv\\
Con \rightarrow (Con \wedge (A \rightarrow B))& \equiv\\
\neg Con \vee (Con \wedge (A \rightarrow B))& \equiv\\
\overline{Con} \vee (Con \wedge (A \rightarrow B))& \\
\end{array}\]

\noindent Variable quantification need not be taken into account
here because each variable occurring in $Con$ must be existentially
quantified in order for the rule to be applied to it. Hence the
quantification of those variables remain unchanged in the two
resulting disjuncts.

\medskip
\noindent \textbf{R7 - Constraint solving}. This rules replaces a
set \{\sequenza{Con}{k}\} of c-conjuncts in $N$ by $\bot$ in $N'$,
provided the constraint solver
evaluates them as unsatisfiable. By the assumption that the
constraint solver is sound and complete, the rule is obviously
equivalence preserving.

\medskip
\noindent \textbf{R8 - Equality rewriting in atoms} and \textbf{R9 -
Equality rewriting in implications}. These rules are directly
borrowed from the Martelli-Montanari unification algorithm. The
equivalence preserving is proven by the soundness of this algorithm
\cite{unification}.

\medskip
\noindent \textbf{R10 - Substitution in atoms} and \textbf{R11 -
Substitution in implications}. These rules simply propagate an
equality either to the whole node or to the implication in which it
occurs. Again they are obviously equivalence preserving rules.

\medskip
\noindent \textbf{R12 - Case Analysis for equality}. The equivalence
preservation of this rule requires some carefulness due to the
quantification of the variables involved. First of all note that
if no variable in the {\bf Given} formula is universally quantified
the proof is trivial. For simplicity we provide the full proof for
the case in which the {\bf Given} formula contains only one
universally quantified variable and no other existentially
quantified variables except $X$. The proof can be then easily
adapted to the general case. With this simplification, we need to
prove that the following two formulae are equivalent (where implicit
quantifications are made explicit).

\medskip
\noindent$\begin{array}{lll}

 \textbf{F1} \quad & \exists X \: \forall Y (  ( X = t \wedge B) \rightarrow H) & \\[2pt]
 \textbf{F2} \quad & [\exists X, Y  (X = t \wedge (B \rightarrow H))] \vee [\exists X \forall Y (X = t\rightarrow \bot )] & \\[2pt]

\end{array}$

\noindent We do a {\it proof by cases}, using the following two
(complementary) hypotheses:

\medskip
$\begin{array}{ll}
 \textbf{Hyp1} \qquad & \neg \exists X \exists Y  (X = t). \\[2pt]
 \textbf{Hyp2} \qquad & \exists X \exists Y  (X = t)\\[2pt]
\end{array}$

\noindent The equivalence under \textbf{Hyp1} is trivial.

Assume \textbf{Hyp2} holds. Let $s$ be a ground value for $X$ such
that

\medskip
$\exists Y  (s = t)$.

\medskip \noindent
and let $\vartheta$ be the ground substitution for $X$ and $Y$ such
that $X\vartheta = s$ and $(X=t)\vartheta$. Note that, by CET,
given $s$ such a ground substitution is unique. Consider now the
formulae obtained from \textbf{F1} and \textbf{F2} by substituting
$X$ by $s$

\medskip
$\begin{array}{ll@{\quad}l}

 \textbf{F1(s)} \qquad & \forall Y ((s = t \wedge B) \rightarrow H) & \\[2pt]
 \textbf{F2(s)} \qquad & [\exists Y  (s= t \wedge (B \rightarrow H))] \vee [\forall Y (s = t \rightarrow \bot )]

\end{array}$

\medskip
\noindent It is not difficult to see that \textbf{F1(s)} is
equivalent to

\medskip
$(B \rightarrow H) \vartheta$

\medskip \noindent
since for any ground instantiation of $Y$ other than $Y\vartheta$
the implication $((s = t \wedge B) \rightarrow H)$ is trivially
true.

\medskip \noindent
Consider now \textbf{F2(s)}. The second disjunct is false by
\textbf{Hyp2} whereas the first disjunct is clearly equivalent to
$(B \rightarrow H) \vartheta$ due to the uniqueness of $\vartheta$.

\medskip
\noindent \textbf{R13 - Negation rewriting}. This rule uses common
logical equivalences:

\[\begin{array}{l@{\quad}l}

((A \rightarrow \bot) \wedge B) \rightarrow H& \equiv\\
B \rightarrow \neg(A \rightarrow \bot) \vee H & \equiv\\
B \rightarrow \neg(\neg A \vee \bot) \vee H & \equiv\\
B \rightarrow (A \wedge \top) \vee H & \equiv\\
B \rightarrow A \vee H & \\
\end{array}\]

\medskip
\noindent \textbf{R14, R15, R16, R17 - Logical simplification \#1 -
\#4} rules. All the four simplification rules are again obviously
equivalence preserving rules as they use common logical
equivalences.

\medskip
\noindent \textbf{R18 - Dynamic Allowedness}. This rule does not
change the elements of a node $N$. Hence, given that $N' = N$,
ignoring the marking, the equivalence preservation is proven.
\end{proof}

\begin{proof}[Proof of Corollary \ref{coroleq}]
\noindent The proof is an immediate consequence of Proposition
\ref{lemmaeq}, because for any CIFF formula $F'$ obtained from $F$
through the application of a CIFF proof rule $\phi$ on a node $N$,
we have that

\[F' = F - \{ N \} \cup \Nodes\]

\noindent where $\Nodes$ is the set of successor nodes of $N$ with
respect to $\phi$.
\end{proof}


\begin{proof}[Proof of Theorem \ref{theo:soundness}]
\noindent Let us consider a CIFF successful node $N$. By definition
of CIFF extracted answer, the node $N$ from which \answerciff{} is
extracted, is a conjunction of the form

\[ \Delta \wedge \Gamma \wedge E \wedge DE \wedge Rest \]

\noindent where $C = \answerC$ and $Rest$ is a conjunction of CIFF
conjuncts.

\noindent Propositions \ref{lemmasub1} and \ref{lemmasub2} ensure
the existence of a ground substitution $\overline{\sigma}$ such
that:

 \[\Delta\overline{\sigma} \models_{3(\Re)} \Delta \cup
 \answerciffCcup. \]

\noindent Let {\cal X} the set of variables occurring in $Q$ and let
$\theta$ the restriction of $\overline{\sigma}$ over the variables
in {\cal X}.

\noindent Let $\gamma$ be a ground substitution for all the
variables occurring in $Q\theta$. Let $\sigma = \theta\gamma$. It is
straightforward that

$\Delta\theta\gamma \models_{3(\Re)} \Delta \cup
 \answerciffCcup$

\noindent as the substitution $\gamma$ does not involve any variable
in $\Delta \cup \answerciffCcup$.

%

\medskip
\noindent To prove that \AnswerC{} is an abductive answer with
constraint, we need that:

\begin{enumerate}
  \item there exists a ground substitution $\sigma'$ for the variables occurring in $\Gamma\sigma$ such that $\sigma' \models_{\Re} \Gamma\sigma$
and
  \item for each ground substitution $\sigma'$ for the variables occurring in $\Gamma\sigma$ such that $\sigma' \models_{\Re} \Gamma\sigma$,
   there exists a ground substitution $\sigma''$ for the variables
occurring in $Q \cup \Delta \cup \Gamma$, with $\sigma\sigma'
\subseteq \sigma''$, such that:

\begin{itemize}

    \item $\Prog \cup \Delta\sigma'' \models_{LP(\Re)} Q\sigma''$ and
    \item $\Prog \cup \Delta\sigma'' \models_{LP(\Re)} IC.$
\end{itemize}
\end{enumerate}

\noindent Again, Propositions \ref{lemmasub1} and \ref{lemmasub2}
ensure that

\begin{itemize}
  \item there exists a ground substitution $\sigma'$ for the variables occurring
in $\Gamma\sigma$ such that $\sigma' \models_{\Re} \Gamma\sigma$ and
such that, for each ground substitution $\sigma'$ and
  \item for each ground substitution $\sigma'$ for the variables occurring in $\Gamma\sigma$ such that $\sigma' \models_{\Re} \Gamma\sigma$,
   there exists a ground substitution $\sigma''$ for the variables
occurring in $Q \cup \Delta \cup \Gamma$, with $\sigma\sigma'
\subseteq \sigma''$, such that:

    \[ \Delta\sigma'' \models_{3(\Re)} \Delta \cup
 \answerciffCcup \qquad (+) \]
\end{itemize}

\noindent If we prove that $\Delta\sigma'' \models_{3(\Re)} Rest$,
we have that

\[P \cup \Delta\sigma'' \models_{3(\Re)} N. \qquad(*)\]

\noindent From this, by induction and by Proposition \ref{lemmaeq},
we will obtain

\begin{itemize}
  \item $P \cup \Delta\sigma'' \models_{3(\Re)} Q\sigma''$, and
  \item $P \cup \Delta\sigma'' \models_{3(\Re)} IC$,
\end{itemize}

\noindent thus proving that \answerciff{} is an abductive answer
with constraints to $Q$ with respect to \AbdCprog.

\medskip

\noindent We now prove $(*)$. It is obvious that:

\[ P \cup \Delta\sigma'' \models_{3(\Re)} \Delta \cup \answerciffCcup \]

\noindent by $(+)$ above. We need to show that:

\[ P \cup \Delta\sigma'' \models_{3(\Re)} Rest. \]

\noindent Let us consider the structure of $Rest$. Due to the
exhaustive application of CIFF proof rules, a CIFF conjunct in
$Rest$ cannot be any of the following:

\begin{itemize}
\item  a disjunction (due to the exhaustive application of
\textbf{Splitting});
\item a defined atom (due to the exhaustive application of
\textbf{Unfolding atoms});
\item  either $\top$ or $\bot$ (due to the exhaustive application of
\textbf{Logical simplification (\#1 - \#4)} and the fact that $N$ is
not a failure node, respectively);
\item  an implication whose body contains a defined atom (due to the exhaustive application of
\textbf{Unfolding in implications});
\item  an implication with a negative literal in the
body (due to the exhaustive application of \textbf{Negation
rewriting});
\item  an implication with $\top$ or $\bot$ in the body
(due to the exhaustive application of \textbf{Logical simplification
(\#1 - \#4)});
\item  an implication with only equalities or constraint atoms in the body
(due to the exhaustive application of \textbf{Case analysis for
equalities}, \textbf{Case analysis for constraints},
\textbf{Substitution in implications} and \textbf{Dynamic
Allowedness}).

\end{itemize}

\noindent Thus, each CIFF conjunct in $Rest$ is an implication whose
body contains at least an abducible atom. We denote as $A_a
\subseteq \Delta$ the set of abducible atoms in $\Delta$ whose
predicate is $a$. Consider an implication $I \in Rest$ of the form
$a(\vec{t}) \wedge B \rightarrow H$ where $a$ is an abducible
predicate and $\vec{t}$ may contain universally quantified
variables.

\noindent Either $A_a = \oslash$ or not. If $A_a = \oslash$ then it
trivially holds that $P \cup \Delta\sigma'' \models_{3(\Re)} I$
because the body of $I$ falsified.

\noindent The case $A_a \neq \oslash$ is more interesting. Assume
$A_a = a(\vec{s}_1), \ldots, a(\vec{s}_k)$. Due to the fact that $a$
has no definition in $P$, $a(\vec{s}_1)\sigma'', \ldots,
a(\vec{s}_k)\sigma''$ represent all and only the instances of
$a(\vec{t})$ which are entailed by $P \cup \Delta\sigma''$ with
respect to the three-valued completion semantics.

\noindent Hence, if $\vec{t} = \vec{s}\sigma''$, where $\vec{s}$ is
such that $a(\vec{s})\sigma'' \not\in A_a$, it trivially holds that
$P \cup \Delta\sigma'' \models_{3(\Re)} I$, because the body of $I$
falsified.

 \noindent Consider now the case $\vec{t} = \vec{s}\sigma''$, where $\vec{s}$ is such
that $a(\vec{s})\sigma'' \in A_a$. Because $N$ is a CIFF successful
node, \textbf{Propagation} has been exhaustively applied in the CIFF
branch \Branch{} whose leaf node is $N$. This means that for each
$a(\vec{s}_i)\sigma'' \in A_a$, an implication $I'$ of the form

\[ \vec{t} = \vec{s}_i\sigma'' \wedge B \rightarrow H \]

\noindent occurs in at least a node $N_i \in {\cal B}$ (otherwise
\textbf{Propagation} is still applicable and $N$ is not a successful
node). Then, if $B$ of the body does not contain other abducibles,
the implication $I'$
is not in $Rest$ and has been reduced to a conjunction in $N$. 

\noindent Otherwise, if $B$ contains another abducible atom, the
process is applied again on it. Because a successful branch is
finite, the proof is obtained by induction on the number of
abducible atoms in $B$.

\medskip
\noindent Hence, it holds that:

\[P \cup \Delta\sigma'' \models_{3(\Re)} Rest \]

\noindent and

\[P \cup \Delta\sigma'' \models_{3(\Re)} N \]

\noindent Let us consider the CIFF branch \Branch{} whose leaf node
is $N$, i.e. the branch $\Branch = N_1 = Q \wedge IC, N_2, \ldots,
N_l = N$ with $l \geq 1$. If we prove that for each pair of nodes
$N_i$ and $ N_{i+1}$ belonging to $\cal{B}$ it holds that if

\[P \cup \Delta\sigma'' \models_{3(\Re)} N_{i+1}\]

\noindent then

\[P \cup \Delta\sigma'' \models_{3(\Re)} N_i\]

\noindent we have, by induction, that

\[P \cup \Delta\sigma'' \models_{3(\Re)} Q\sigma'' \wedge IC\]

\noindent Suppose $P \cup \Delta\sigma'' \models_{3(\Re)} N_{i+1}$,
for some $i$. Due to the definition of CIFF branch, each node
$N_{i+1} \in \cal{B}$ is one of the successor nodes of $N_i$. If
$N_{i+1}$ is obtained by $N_i$ by applying a CIFF proof rule
distinct from the \textbf{Splitting} rule, if follows immediately
that

\[ P \cup \Delta\sigma'' \models_{3(\Re)} N_i \]

\noindent given that $N_{i+1}$ is the only successor node of $N_i$
and thus, from Proposition \ref{lemmaeq}, we have that $N_i \equiv
N_{i+1}$. If the \textbf{Splitting} rule has been applied, however,
then the node $N_i$ is of the form

\[ RestNode \wedge (D_1 \vee \ldots \vee D_n)\]

\noindent and $N_{i+1}$ is of the form

\begin{center}
$(RestNode \wedge D_i)$ \qquad for some $i \in [1,n]$.
\end{center}

\noindent It is obvious that the latter formula entails the former.

\medskip
\noindent Summarizing, we have that

\[P \cup \Delta\sigma'' \models_{3(\Re)} Q\sigma'' \wedge IC\]

\noindent which implies that

     $P \cup \Delta\sigma'' \models_{3(\Re)} Q\sigma''$, and

     $P \cup \Delta\sigma'' \models_{3(\Re)} IC$.
\end{proof}

\begin{proof*}[Proof of Theorem \ref{theo:soundnessfail}]
\noindent From the definition of failure CIFF derivation, \Der{}
 is a derivation starting with $Q \cup IC$ and such that all its leaf nodes are CIFF failure nodes which are equivalent
to $\bot$.

\noindent Hence, due to Corollary \ref{coroleq} and the transitivity
of the equivalence, it follows immediately that:

 \[  P \cup IC \models_{3(\Re)} (Q \wedge IC) \leftrightarrow \bot \]

\noindent Because $IC$ occurs in both the left and the right hand
side of the statement, we have that

\[  P  \cup IC \models_{3(\Re)} Q \leftrightarrow \bot \]

\noindent and thus

 \[  P  \cup IC \models_{3(\Re)} \neg Q. \mathproofbox \]
\end{proof*}

\noindent The proof of Lemma \ref{lemma:static_allowedness} requires
some auxiliary definition and result given in the sequel

\begin{definition}
An atom is a {\em pure} constraint atom if is either a
constraint atom or it is an equality $t=s$ where either
$t$ or $s$ are non-Herbrand terms.
\end{definition}

\noindent For example the equality $X=3$ is a {\em pure}
constraint atom whereas the equality $X=a$ is not.

\begin{definition}[Statically allowed implication]\label{def:static_allowed_implication}
An implication of the form $B \rightarrow H$ is {\em statically
allowed} if and only if:

\begin{itemize}
  \item each universally quantified variable occurring in $H$ occurs
  also in $B$;
  \item each universally quantified variable occurring in a negative literal or in
  a pure constraint atom in $B$, occurs also in an atomic non-constraint atom in $B$;
  \item if a universally quantified variable in $B$ occurs only in an equality
  $t=s$ of $B$ then either $t$ or $s$ do not contain universally quantified variables.
\end{itemize}

\end{definition}

\begin{lemma}[Static allowed implications lemma]
\label{lemma:static_allowed_implications} Let \AbdCprog{} be an
abductive logic program with constraints such that the corresponding
CIFF framework \abdciffframe{} and the query $Q$ are both CIFF
statically allowed. Let \Der{} be a CIFF derivation with respect to
\abdciffframe{} and $Q$. Let $F_i$ be a CIFF formula in \Der{} and
let $N$ be a CIFF node in $F_i$ such that each implication (as a
CIFF conjunct) in $N$ is statically allowed. Then, for each CIFF
proof rule $\phi$ such that

 \ruleappargs{F_i}{$N,\Cs$}{$\phi$}{F_{i+1}},

\noindent each node $N'$ in the set of CIFF successor nodes \Nodes{}
of $N$ in \Der{} is such that each implication (as a CIFF conjunct)
in $N'$ is statically allowed.

\end{lemma}

\begin{proof}[Proof of Lemma \ref{lemma:static_allowed_implications}]
\noindent We need to prove that each implication $I$ of the form $B
\rightarrow H$ in each successor node $N'$ of $N$ is statically
allowed.

\noindent For all CIFF proof rules but (R1), (R2), (R3), (R9),
(R11), (R12) and (R13) the proof is trivial.

\medskip
\noindent \textbf{Unfolding atoms (R1)}. This rule resolves an atom
$p(\vec{t})$ with its iff-definition $[p(\vec{X})\equivalent
D_1\vee\cdots\vee D_n]\in\Th$. New implications can arise from
negative literals (rewritten in implicative form) in some disjunct
$D_i$ ($i \in [1,n]$). However, by assumption, $Th$ is statically
allowed and thus each universally quantified variable $V$ occurring
in a negative literal occurs elsewhere in a non-equality,
non-constraint atom in the same disjunct. Hence any such newly
introduced implication is statically allowed.

\medskip
\noindent \textbf{Unfolding within implications (R2)}. This rule
resolves an atom $p(\vec{t})$ in the body of an implication with its
iff-definition $[p(\vec{X})\equivalent D_1\vee\cdots\vee
D_n]\in\Th$, producing $n$ new implications $I_1, \ldots, I_n$ in
the successor node of $N$. As for the previous case, since $Th$ is
statically allowed, each universally quantified variable $V$
occurring in a disjunct $D_i$ ($i \in [1,n]$) occurs elsewhere in a
non-equality, non-constraint atom in the same disjunct. Hence each
$I_i$ ($i \in [1,n]$) is a statically allowed implication.

\medskip
\noindent \textbf{Propagation (R3)}. This rule resolves an atom
$p(\vec{t})$ in the body of an implication $I$ with an atom
$p(\vec{s})$ as a CIFF conjunct in $N$, adding a new implication
$I'$ in the successor node of $N$, where $p(\vec{t})$ is replaced by
$\vec{t}=\vec{s}$. By definition, all the variables in $\vec{s}$ are
existentially quantified, hence the newly introduced implication is
statically allowed.

\medskip
\noindent \textbf{Equality rewriting in implications (R9)}. This
rule handles an implication $I$ of the form $(t_1 = t_2 \wedge
B)\imp H$, replacing it with an implication $I'$ of the form
$(({\cal E}(t_1 = t_2) \wedge B)\imp H$ in the successor node $N'$
of $N$. Assume that $I'$ is not a statically allowed implication.
There are two cases:
\begin{itemize}
\item a universally quantified variable $V$ in $H$ occurred in $B$
only in the equality $t_1=t_2$ and the application of ${\cal E}(t_1,
t_2)$ has eliminated $V$. This can never happen since, being $I$
statically allowed, cases (4) and (5) in the definition of ${\cal
E}$ do not apply;
\item a universally quantified variable $V$ occurring only in $t_1=t_2$ still
occurs only in an equality $t'=s'$ introduced by the application of
${\cal E}(t_1=t_2)$, and both $t'$ and $s'$ contain universally
quantified variables. This can not happen either, since $t'$ is a
subterm of $t_1$, $s'$ is a subterm of $t_2$ and either $t_1$ or
$t_2$ do not contain universally quantified variables by the
hypothesis that $I$ is statically allowed.
\end{itemize}

\medskip
\noindent \textbf{Substitution in implications (R11)}. This rule
handles an implication $I$ of the form $(X = t \wedge B)\imp H$
(where $X$ is universally quantified and $X$ does not occur in $t$),
replacing it with an implication $I'$ of the form $(B \imp H)[X/t]$
in the successor node $N'$ of $N$. Since $I$ is statically allowed
and $I'$ contains one less universally quantified variable with
respect to $I$, $I'$ is also statically allowed.

\medskip
\noindent \textbf{Case analysis for equalities (R12)}. This rule
handles an implication $I$ of the form $(X=t\wedge B)\imp H$, (where
$X$ is existentially quantified) replacing it with a disjunctive
node of the form $[X=t \wedge (B \imp H)] \vee [X=t\imp\bot]$ (where
all the variables in $t$ in the first disjunct become existentially
quantified) in the successor node $N'$ of $N$. Being $X$
existentially quantified, the implication $X=t\imp\bot$ in the
second disjunct is statically allowed. Moreover, due to the fact
that all the variables in $t$ become existentially quantified in the
first disjunct, also $B \imp H$ is statically allowed because it
contains less universally quantified variables than $I$ which is, by
assumption, statically allowed.

\medskip
\noindent \textbf{Negation rewriting (R13)}. This rule handles an
implication $I$ of the form $((A\imp\bot)\wedge B)\imp H$ ,
replacing it with an implication $I'$ of the form $B\imp (A\vee H)$
in the successor node $N'$ of $N$. Being $I$ statically allowed, for
each variable $V$ occurring in $A$, $V$ must also occur in a
non-equality, non-constraint atom in $B$ and thus also $I'$ is
statically allowed because each variable in $(A\vee H)$ occurs also
in a non-equality, non-constraint atom in $B$.

\end{proof}

\begin{corollary}
\label{corollario} Let \AbdCprog{} be an abductive logic program
with constraints such that the corresponding CIFF framework
\abdciffframe{} and the query $Q$ are both CIFF statically allowed.
Let \Der{} be a CIFF derivation with respect to \abdciffframe{} and
$Q$. Then each implication occurring in \Der{} is a statically
allowed implication.
\end{corollary}

\begin{proof}
Any implication in the initial node of \Der{} is statically allowed
since the \abdciffframe{} and the query $Q$ are both CIFF statically
allowed by hypothesis. The result then follows directly from Lemma
\ref{lemma:static_allowed_implications}.
\end{proof}


\begin{proof}[Proof of Lemma \ref{lemma:static_allowedness}]
\noindent We prove the Lemma by contradiction. Assume that there
exists a CIFF derivation such that \textbf{R18 - Dynamic
allowedness} is selected. By definition of the \textbf{Dynamic
allowedness} rule, an implication of form $B \rightarrow H$  is
selected such that:
\begin{itemize}
\item[(i)]either $B$ is $\top$, or
\item[(ii)] $B$ contains constraint atoms only
\end{itemize}
and
\begin{itemize}
\item[(iii)] no other rule applies to the implication.
\end{itemize}

\noindent Due to the definition of the CIFF proof rules, (i), (ii)
and (iii) above imply that
\begin{itemize}
  \item[(iv)] either $B$ is $\top$ and $H$ contains universally quantified
  variables, or
  \item[(v)] $B$ contains constraint atoms only,
  each constraint atom in $B$ contains universally quantified
variables, and each equality atom in $B$ is a pure constraint atom.
\end{itemize}

\medskip
\noindent Note, in particular, that equalities in $B$ are pure
constraint atoms since otherwise \textbf{R9}, \textbf{R11} or
\textbf{R12} would be applicable. In both cases (iv) and (v) the
implication is not a statically allowed implication, contradicting
Corollary \ref{corollario}.
\end{proof}

\begin{proof}[Proof of Theorem \ref{theo:IFFcompleteness}]
\noindent By assumption, both \abdciffframe{} and $Q$ do not contain
constraint atoms. This means that both the CIFF framework
\abdciffframe{} and the CIFF query $Q$ are also an IFF framework and
an IFF query respectively. Moreover, the CIFF proof rules are a
superset of the IFF proof rules. Directly from the same assumption
\textbf{Case analysis for constraints} and \textbf{Constraint
solving} (which are all the CIFF rules managing c-atoms) can never
be applied in any derivation $\overline{\Der}${} for $Q$ with
respect to \abdciffframe.

\noindent Moreover, the fact that both \abdciffframe{} and $Q$ are
IFF allowed ensures that they are also CIFF statically allowed. This
is trivial because an IFF allowed query is defined exactly as a CIFF
statically allowed query and the notion of CIFF static allowedness
and the notion of IFF allowedness for, respectively, a CIFF and an
IFF framework, differ only for the CIFF static allowedness
conditions over constraint atoms. As \abdciffframe{} does not
contain constraint atoms, the two notions for \abdciffframe{}
coincide. Hence \abdciffframe{} is also a CIFF statically allowed
framework and thus, for Lemma \ref{lemma:static_allowedness},
\textbf{Dynamic allowedness} is never applied. This means that any
derivation $\overline{\Der}${} for $Q$ with respect to
\abdciffframe{} is an IFF derivation and thus, we can apply directly
the completeness result stated in \cite{iff97}.
\end{proof}

\begin{proof}[Proof of Theorem \ref{theo:completeness}]

\begin{enumerate}
\item It is easy to see that

 \[  P  \cup IC \models_{3(\Re)} \neg Q  \]

 \noindent is equivalent to:
\[  P  \cup IC \models_{3(\Re)} Q \leftrightarrow \bot \]

 \noindent Because $IC$ occurs in the left hand side of the statement, the above statement is
equivalent to:

 \[  (*) \qquad \qquad P \cup IC \models_{3(\Re)} (Q \wedge IC) \leftrightarrow \bot \]

\noindent Assume that there exists a CIFF successful branch in
\Der{} and let $Ans$ be the corresponding CIFF extracted answer. Due
to the equivalence preservation of CIFF rules (Proposition
\ref{coroleq}) and the transitivity of the equivalence, we have that

\[  P \cup IC \models_{3(\Re)} (Q \wedge IC) \leftrightarrow (\bot \vee Ans) \]

\noindent which clearly contradicts the above statement $(*)$ being
$Ans$ distinct from $\bot$ due to the soundness of CIFF.

\item Assume that all the branches in
\Der{} are failure branches. Due to the equivalence preservation of
CIFF rules (Proposition \ref{coroleq}) and the transitivity of the
equivalence, we have that

\[  P \cup IC \models_{3(\Re)} (Q \wedge IC) \leftrightarrow \bot \]

\noindent which is equivalent to

\[  P  \cup IC \models_{3(\Re)} Q \leftrightarrow \bot \]

\noindent and to

 \[  P  \cup IC \models_{3(\Re)} \neg Q  \]

\noindent which clearly contradicts that

 \[  P  \cup IC \not\models_{3(\Re)} \neg Q.  \]
\end{enumerate}
\end{proof}

\begin{proof}[Proof of Theorem \ref{theo:static_completeness}]
By Lemma \ref{lemma:static_allowedness} we have that, given a CIFF
derivation \Der{} with respect to \abdciffframe{} and $Q$, \Der{}
does not contain undefined branches. This is because the
\textbf{Dynamic Allowedness} rule is never applied in \Der{} and
this is the only rule which gives rise to an undefined node. Due to
the assumption that \Der{} is finite, we have that all the final
nodes in \Der{} are either successful or failure CIFF nodes. Hence
Theorem \ref{theo:completeness} can be applied to \abdciffframe{}
and $Q$, thus proving the statement.
\end{proof}

\end{document}